\documentclass[10pt,twocolumn,letterpaper]{article}

\usepackage{iccv}
\usepackage{times}
\usepackage{epsfig}
\usepackage{graphicx}
\usepackage{amsmath}
\usepackage{amssymb}

\usepackage[numbers,sort,compress]{natbib}
\usepackage{soul}
\usepackage[dvipsnames]{xcolor}
\usepackage{xspace}
\usepackage[normalem]{ulem}

\usepackage{tabularx,ragged2e,booktabs}
\usepackage{multirow}
\usepackage[percent]{overpic}

\usepackage[figuresleft]{rotating}

\usepackage{shadowtext}
\usepackage[export]{adjustbox}
\usepackage{pifont}
\newcommand{\cmark}{\ding{51}}%
\newcommand{\xmark}{\ding{55}}%

\usepackage{color, colortbl}
\definecolor{Gray}{gray}{0.9}

\usepackage{enumitem}

\usepackage[heightadjust=all,valign=c]{floatrow}
\newfloatcommand{capbtabbox}{table}[][\FBwidth]

\usepackage[outline]{contour}
\usepackage{mathtools}
\usepackage{bbm}

\usepackage{sg-macros}

\makeatletter
\renewcommand{\paragraph}{%
  \@startsection{paragraph}{4}%
  {\z@}{1.05ex \@plus 1ex \@minus .2ex}{-1em}%
  {\normalfont\normalsize\bfseries}%
}
\makeatother
\usepackage[pagebackref=true,breaklinks=true,colorlinks,bookmarks=false]{hyperref}

\newcommand{\dstname}{CIRR\xspace}
\newcommand{\dstnamefull}{Compose Image Retrieval on Real-life images\xspace}
\newcommand{\modelnamefull}{Composed Image Retrieval using Pretrained LANguage Transformers\xspace}
\newcommand{\modelname}{CIRPLANT\xspace}
\newcommand{\dsturl}{\url{https://cuberick-orion.github.io/CIRR/}\xspace}

\iccvfinalcopy 

\begin{document}

\title{Image Retrieval on Real-life Images with Pre-trained \\Vision-and-Language Models}

\author{Zheyuan Liu$^{1}$ \quad Cristian Rodriguez-Opazo$^{2}$ \quad Damien Teney$^{2,3}$ \quad Stephen Gould$^{1}$ \\
$^{1}$Australian National University\\
$^{2}$Australian Institute for Machine Learning, University of Adelaide\quad
$^{3}$Idiap Research Institute \\
{\tt\small \{zheyuan.liu, stephen.gould\}@anu.edu.au} \\
{\tt\small cristian.rodriguezopazo@adelaide.edu.au, damien.teney@idiap.ch}
}

\maketitle

\ificcvfinal\thispagestyle{empty}\fi

\begin{abstract}
   We extend the task of composed image retrieval, where an input query consists of an image and short textual description of how to modify the image.
   Existing methods have only been applied to non-complex images within narrow domains, such as fashion products,
   thereby limiting the scope of study on in-depth visual reasoning in rich image and language contexts.
   To address this issue, we collect the \dstnamefull (\dstname) dataset, which consists of over 36,000 pairs of crowd-sourced, open-domain images with human-generated modifying text.
   To extend current methods to the open-domain, we propose \modelname, a transformer based model that leverages rich pre-trained vision-and-language (V\&L) knowledge for modifying visual features conditioned on natural language.
   Retrieval is then done by nearest neighbor lookup on the modified features.
   We demonstrate that with a relatively simple architecture, \modelname outperforms existing methods on open-domain images, while matching state-of-the-art accuracy on the existing narrow datasets, such as fashion.
   Together with the release of \dstname, we believe this work will inspire further research on composed image retrieval.
   Our dataset, code and pre-trained models are available at \dsturl.
\end{abstract}

\section{Introduction} \label{sec:intro}
We study the task of \emph{composed image retrieval}, that is, finding an image from a large corpus that best matches a user query provided as an image-language pair. Unlike traditional content-based~\cite{10.1145/500141.500159_cbir} or text-based~\cite{zhang_tbir,6126478_tbir} image retrieval where a single modality is used to describe the target image, composed image retrieval involves both visual and textual modalities to specify the user's intent. For humans the advantage of a bi-modal query is clear: some concepts and attributes are more succinctly described visually, others through language. By cross-referencing the two modalities, a reference image can capture the general gist of a scene, while the text can specify finer details. 
The challenge is the inherent ambiguity in knowing what information is important (typically one object of interest in the scene) and what can be ignored (\eg, the background and other irrelevant objects). However, existing datasets for this task fall short of allowing us to adequately study this problem.

\begin{figure}[t]
  \begin{center}
    \includegraphics[trim={0 5pt 0 0},clip, width=0.99\linewidth]{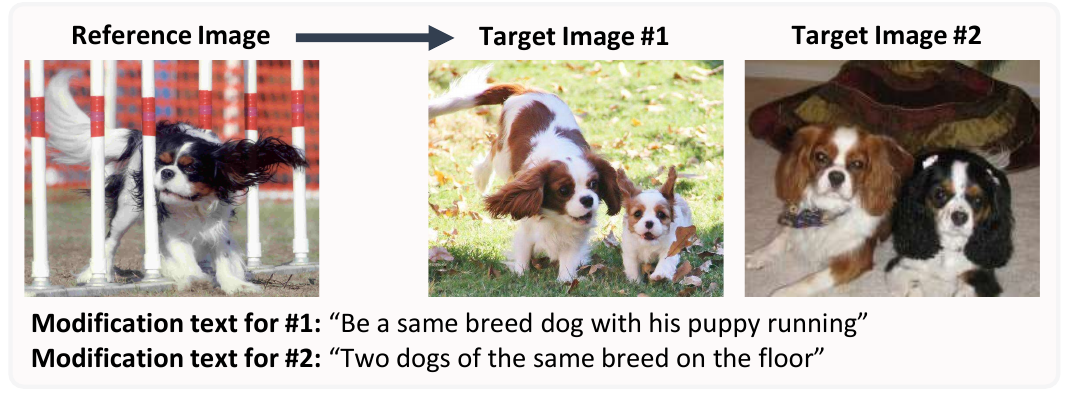}
  \end{center}
   \caption{Example of composed image retrieval from the proposed \dstname dataset. The input is composed of a reference image and a modifying text, to which the model must find a close match. A major challenge is the inherent ambiguity and underspecification of visual aspects to be preserved or modified. Our dataset includes open-domain images with rich contexts to facilitate the study of such challenge.}
   \label{fig:intro-0}
\end{figure}

Consider the example in \figref{fig:intro-0}. 
Real-life images usually contain rich object interactions on various scales. In each case, to readily identify the relevant aspects to keep or change and pay less attention elsewhere (\eg, the color of the dog's fur and background objects), a model must develop in-depth visual reasoning ability and infer implicit human agreements within both the visual and language contexts.
However, existing datasets are constrained to domains such as fashion products~\cite{fashioniq,han2017automatic_fashion200k,10.5555/1886063.1886114_shoes} or synthetic objects~\cite{Vo_2019_tirg} with relatively simple image contents.
We argue that the current datasets are insufficient for exploring the unique research opportunity mentioned above.

Motivated by this problem, we collect the \dstnamefull (\dstname) dataset. It is based on the open-domain collection of real images from NLVR$^2$~\cite{Suhr_2019_nlvr2}, for which we collected rich, high-quality annotations that aim to tease out the important aspects of the reference image and textual description for a given query.

Compared with existing datasets, \dstname places more emphasis on distinguishing between visually similar images, which provides a greater challenge, as well as a chance for studying fine-grained vision-and-language (V\&L) reasoning in composed image retrieval. 
Our dataset also allows for evaluation on fully labeled subsets, which addresses a shortcoming of existing datasets that are not fully labeled and therefore contain multiple false-negatives (as unlabeled images are considered negative).

Meanwhile, we propose \modelnamefull (\modelname), which extends current methods into open-domain images by leveraging the knowledge of large-scale V\&L pre-trained (VLP) model~\cite{oscar}. 
Although the advantages of such pre-trained models have been validated in many visiolinguistic tasks~\cite{oscar,vilbert,chen2020uniter}, to the best of our knowledge, none have been applied to composed image retrieval. We conjecture one of the reasons being the existing domain-specific datasets cannot greatly benefit from the pre-training, which uses more complex, open-world images. 
Moreover, to adopt the VLP models for fine-tuning, most of the downstream tasks are formulated as classification tasks~\cite{oscar,chen2020uniter}. For composed image retrieval, it requires taking as input both the reference and target images. However, this greatly raises the computational overhead for retrieval, as the model needs to exhaustively assess each input query paired with each candidate target before yielding the one with the highest prediction score. 
Instead, we propose to preserve the conventional metric learning pipeline, where the input queries are jointly embedded using the VLP model and later compared with features of candidate images through $\ell_2$-norm distance. 
Specifically, our design maintains the same objective of ``language-conditioned image feature modification'' as previous work~\cite{Vo_2019_tirg,chen2020image_val,dodds2020modality_maaf}, while manages to utilize the pre-trained V\&L knowledge in large-scale models.
We demonstrate that our proposed model reaches state-of-the-art on the existing fashion dataset while outperforming current methods on \dstname.

\section{Related Work}\label{sec:related_work}

\paragraph{Image retrieval.}
Existing work on image retrieval using deep learning can be categorized by the type of queries considered.
Content-based Image Retrieval (CBIR) refers to the use of image-only queries for product search~\cite{liuLQWTcvpr16DeepFashion}, face recognition~\cite{schroff2015facenet,masi2018deep}, etc.
This setup leaves little room for iterative user feedback or refinement.
Other possible modalities to form queries include attributes~\cite{han2017automatic_fashion200k}, natural language~\cite{6126478_tbir,zhang_tbir}, and sketches~\cite{radenovic2018deep}.
These are motivated by a more natural user experience, but require more advanced retrieval mechanisms.
\citet{Vo_2019_tirg} propose \emph{composed image retrieval} that combines visual and text modalities.
Here the query consists of a reference image and short text describing desired differences with this image.
\citet{fashioniq} demonstrate the potential of this setup for the narrow domain of fashion recommendation.

Our work focuses on composed image retrieval in an open-domain setting, \ie, not restricted to fashion products for example. We specifically address the case of distinguishing visually similar images, which requires more in-depth, fine-grained reasoning ablility over both the visual and language modalities.

\paragraph{Compositional learning.} 
The topic of compositional learning has been extensively studied in V\&L tasks including visual question answering (VQA)~\cite{antol2015vqa}, image captioning~\cite{aneja2017convolutional, anderson2018bottom} and video retrieval~\cite{Xu_2019-T2C}.
The aim is to produce learned joint-embedding features that capture the salient information in both visual and text modalities along with their interactions.
For composed image retrieval, \citet{Vo_2019_tirg} first propose a residual-gating mechanism that aims to control variation of the input image features through text.
\citet{9157125_hosseinzadeh} use region-based visual features from R-CNN models~\cite{girshick2015fast_fastrcnn,ren2015faster_fasterrcnn} originally proposed for image captioning~\cite{anderson2018bottom} and VQA~\cite{teney2017tipsandtricks}. 
Recently, \citet{chen2020image_val} use a transformer-based model~\cite{vaswani2017attention_transformer} and inject the text modality at varying depths of the image model. \citet{dodds2020modality_maaf} introduce the concept of modality-agnostic tokens, which they obtain from ``divided'' spatial convolutional features and LSTM hidden states.
In this work, we propose a method that leverages the rich knowledge in VLP models. Our method can modify the input image features based on natural language without the need of developing monolithic architecture on the specific task.

\paragraph{Vision-and-language pre-training.}
The success of pre-trained BERT~\cite{devlin2018bert_bert} inspired numerous attempts on VLP models, including~\cite{chen2020uniter,vilbert,li2019_visualbert,oscar,tan2019lxmert}. The aim is to develop Transformer-based~\cite{vaswani2017attention_transformer} models trained on large-scale image-text triplets to produces V\&L representations applicable to various tasks. The advantage is clear, instead of training monolithic models on task-specific datasets from zero, different V\&L tasks can start with the representations learned from (usually) a considerably larger image-text corpus, and fine-tune on specific tasks.
Motivated by success in other V\&L tasks, we propose to adopt the VLP model on composed image retrieval. 
The key obstacle is to design the architecture to encourage a controlled modification of image features, which, differs greatly from the conventional use cases of such models.

\paragraph{Datasets for composed image retrieval.}
Most existing datasets suitable for composed image retrieval are repurposed from other tasks~\cite{Isola2015DiscoveringSA_mitstates,han2017automatic_fashion200k,Vo_2019_tirg}. Images are paired within classes and textual descriptions of their differences are generated automatically from existing labels. 
These datasets are relatively simple visually and only contain short descriptions with simple language.
CSS~\cite{Vo_2019_tirg} uses the synthetic images of geometric 3D shapes from CLEVR~\cite{clevr}, paired with descriptions generated according to differences in appearance of the objects.
Fashion200k~\cite{han2017automatic_fashion200k} contains approx. 200k images tagged with attributes that can be used to compose text descriptions of differences between images. 
MIT-States~\cite{Isola2015DiscoveringSA_mitstates} contains images of entities in different states each labelled with one noun and one adjective. The adjectives can describe limited differences between images.
More recent works introduced human-generated descriptions.
\citet{guo2018dialog} present annotations for Shoes~\cite{10.5555/1886063.1886114_shoes}, a dataset of 10k footwear images. 
Fashion-IQ~\cite{fashioniq} contains crowd-sourced descriptions of differences between images of fashion products. 
\citet{dodds2020modality_maaf} introduce benchmarks for the Birds-to-Words~\cite{forbes2019neural_birds} and Spot-the-Diff~\cite{jhamtani2018learning_spotthediff} datasets.

In this paper, we introduce a new dataset that addresses current deficiencies.
Our dataset is open-domain and not restricted, \eg, to fashion products~\cite{han2017automatic_fashion200k,fashioniq,10.5555/1886063.1886114_shoes}.
We design a careful collection process to produce high-quality pairs from our diverse collection of images by only associating visually- and semantically-related images.
We also address the issue of false-negative targets, that is, candidate target images that are valid for a certain input query, but not labeled as such.
Previous datasets failed to resolve this issue due to the cost of exhaustively labeling images against every possible query, which is mitigated by our data collection strategy.
Although not used in our current work, the dataset also contains a rich set of auxiliary annotations that clarify ambiguities not addressed in the textual query.

\section{The Proposed Model}\label{model}
In this section, we first briefly introduce the vision-and-language pre-trained (VLP) models, then we discuss our adaptation of it for the task of composed image retrieval. 

\begin{figure*}[t!]
  \begin{center}
    \includegraphics[trim={0pt 1pt 0 6pt},clip, width=0.85\linewidth]{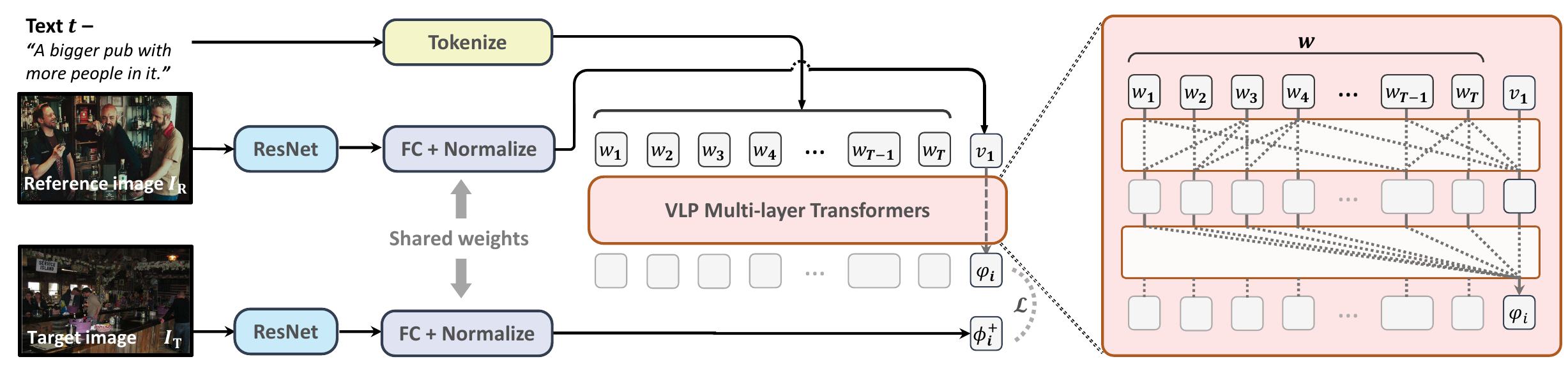}
  \end{center}
  \caption{(Left) Schematic of our model. Given a pair of reference image and text as input, we aim at learning a \textit{modified} image feature of the reference image conditioned on the text, such that it matches the feature of the target image. To compare image features of reference and candidate target images, we extract ResNet features and use a shared FC-layer (with normalization) to project them into the same domain. (Right) Overview of the image-text composition module using vision-and-language pre-trained (VLP) multi-layer transformers. Dashed lines (not fully drawn) represent feature aggregation by attention, which learns a language-conditioned image feature modification.}
  \label{fig:model-0}
  \vspace{-1.0em}
\end{figure*}

\begin{figure*}[!ht]
  \centering
    \includegraphics[trim={0 1pt 0 5pt},clip, width=0.90\linewidth]{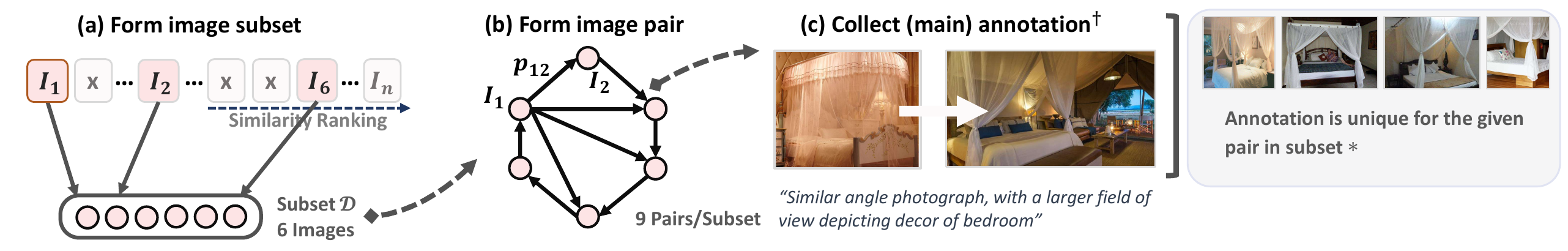}
  \caption{Overview of the data collection process. (a) We demonstrate the construction of an image subset. (b) We illustrate how we choose and form 9 image pairs within one subset, where each arrow suggests the direction from a reference to a target image. (c) $\dagger$ represents Human Tasks with AMT workers. $\ast$ indicates the instruction that mitigates the issue of false-negative.}
  \label{fig:dset_collection_0}
\end{figure*}

\subsection{Vision-and-Language Pre-trained Models}

Contemporary VLP models are inspired by BERT~\cite{devlin2018bert_bert}, which is constructed with multi-layer transformers~\cite{vaswani2017attention_transformer}. 
The model accepts variable-length sequential inputs $\boldsymbol{i}_{\text{VLP}}$, which consist of a concatenation among words in the text sequence(s) $\boldsymbol{w}=\{w_1,\ldots,w_T\}$, regional features from the image $\boldsymbol{v}=\{v_1,\ldots,v_K\}$, and other optional tokens. For instance, in OSCAR~\cite{oscar}, an object label associated with each regional feature is appended to the end as $\boldsymbol{l}=\{l_1,\ldots,l_K\}$.

Within each transformer layer, a multi-head self-attention mechanism is designed to capture the dependencies among the sequential tokens. Layers are stacked hierarchically to attend to the output of the previous layer. Once pre-trained on a large corpus, the final output representations can be used for fine-tuning on arbitrary downstream tasks, where the usage varies depending on the task.

That said, downstream tasks share some common aspects.
Mostly, a classification token \texttt{[CLS]} is inserted at the start of the input text sequence, which aggregates information from the modalities. The final \texttt{[CLS]} output is then used to make predictions, such as for image classification.

\subsection{Adaptation to Composed Image Retrieval}
The task of composed image retrieval can be formally described as finding the target image in a large corpus of images $I_{\text{T}} \in \D$ that best matches a query provided by a reference image-text pair $q = \langle I_{\text{R}}, t\rangle$. 
Our goal is to learn a text-image composition module, which maps a given $\langle I_{\text{R}}, t\rangle$ into the same embedding space as, and close to, the corresponding $I_{\text{T}}$. Intuitively speaking, this requires the composition module to modify $I_{\text{R}}$ conditioned on $t$.

In this work, we employ OSCAR~\cite{oscar}, a recently proposed VLP model with state-of-the-art performance as the composition module to perform the mapping as follows. 

\paragraph{Input sequence.}
We denote the input sequence of OSCAR as $\boldsymbol{i}_{\text{VLP}}=\{\boldsymbol{w},\boldsymbol{v}\}$, where we initialize OSCAR without the optional object label inputs $\boldsymbol{l}$. We then follow Li \etal~\cite{oscar} for processing text sequences, but introduce the following adaptations on image representations.

Rather than including a set of regional features, we pre-process images through an ImageNet pre-trained ResNet~\cite{he2015deep} model and extract features from before the final FC-layer. We then process these features through a (newly) learned FC-layer and $\ell_2$-normalization to give a single image feature $\bv = \{v_1\}$ as the input to OSCAR. This same feature representation is used for the corpus of candidate target images $I'_{\text{T}} \in \D$ as shown in \figref{fig:model-0}.

We choose this relatively simple design for two reasons. First, recent work (\eg,~\cite{hong2020recurrent}) has shown the compatibility between VLP models and non-regional features of images. Second, we hypothesize that using global image features is easier to achieve our goal of modifying $I_{\text{R}}$ conditioned on $t$ so as to closely match $I_{\text{T}}$.

\paragraph{Output token.}
As shown in \figref{fig:model-0}, contrary to typical downstream tasks, we do not use the final representation of the \texttt{[CLS]} token as the text-image joint embedding. Instead, we extract the representation corresponding to the image feature token and treat it as the composed image-text feature.
This resembles the fine-tuning of REF~\cite{li2019_visualbert}, as well as VLN-BERT~\cite{hong2020recurrent}. In both cases, tokens other than \texttt{[CLS]} are used for prediction. 
For composed image retrieval, our design makes sense since the transformer model includes residual connections between input and output tokens. Intuitively, the reference image features are \textit{modified} by aggregating the information from other word tokens to produce the target image features.

\paragraph{Metric learning.}
We use soft triplet-based loss with $\ell_2$-norm distance as in Vo \etal~\cite{Vo_2019_tirg} to bring the composed image-text feature closer to the feature of the target image (positive pair), while pulling apart the features of negative pairs. 
In essence, given the $i$-th positive pair $\langle \varphi_i, \phi^{+}_i \rangle$ and an arbitrary negative $\phi^{-}_{i,j}$ among all negatives $\phi^{-}_{i}$, the loss is computed as:
\begin{equation}
  \mathcal{L} = \log [ 1 + \exp ( \kappa(\varphi_i,\phi^{-}_{i,j}) - \kappa(\varphi_i,\phi^{+}_i) ) ],
\end{equation}
where $\kappa$ is $\ell_2$-norm distance. In training, we randomly sample the negative for each pair and average the loss over all sampled triplets $\langle \varphi_i, \phi^{+}_i, \phi^{-}_{i,j} \rangle$.

\section{The \dstname Dataset}\label{dataset}
 
 Existing datasets for composed image retrieval~\cite{Vo_2019_tirg,fashioniq} contain training and testing examples as triplets $\langle I_{\text{R}}, q, I_{\text{T}} \rangle$ where $q = \langle I_{\text{R}}, t\rangle$ forms the query and $I_{\text{T}}$ is (an example of) the desired target from a large image corpus $\D$.
 However, these existing datasets have two major shortcomings.
 First, they lack the sufficient visual complexity to facilitate the study of one of the major challenges in composed image retrieval, which is the subtle reasoning over what aspects are important and what shall be ignored.
 Second, since the candidate images cannot be extensively labeled for each $\langle I_{\text{R}}, t\rangle$ pair, existing datasets contain many false-negatives. That is, images $I \in \D$ that are valid matches for the query but not labeled as the ground-truth target $I_{\text{T}}$. Indeed, all images in $\D \setminus \{I_{\text{R}}, I_{\text{T}}\}$ are considered as negatives. To circumvent this shortcoming, existing works choose to evaluate models with Recall$@K$ and set $K$ to larger values (\eg, 10, 50~\cite{fashioniq}), thus accounting for the presence of false-negatives. 
 However, the issue persists during training.
 Moreover, by setting larger $K$ values, these methods are essentially trading in their ability for learning detailed text-image modifications.
 
 To mitigate these issues, we introduce the \dstnamefull (\dstname) dataset, which includes over 36,000 annotated query-target pairs, $\langle q = \langle I_{\text{R}}, t \rangle, I_{\text{T}}\rangle$. Unlike existing datasets, we collect the modifying text to distinguish the target from a set of similar images (addressing the problem of false-negatives) and creating challenging examples that require careful consideration of visual and textual cues. Details are as follows.

 \subsection{Data Collection}
 We first form image pairs then collect related annotations by crowd-sourcing.
 The pairs are drawn from subsets of images, as described below. This strategy plays a major role in mitigating the issue of false negatives (see \secref{sec:extended_metric}).
 \figref{fig:dset_collection_0} outlines our data collection procedure.
 
 \paragraph{Image source.} We use the popular NLVR$^2$ dataset for natural language visual reasoning~\cite{Suhr_2019_nlvr2} as our source of images.
 We choose NLVR$^2$ for several reasons. 
 First, it contains images of real-world entities with reasonable complexity in ImageNet-type~\cite{krizhevsky2012imagenet}. 
 Second, the setup of our task requires image in pairs that are similar enough, and NLVR$^2$ is designed to have collections of similar images regarding 1,000 synsets (\eg, acorn, seawall). 
 Also, \citet{Suhr_2019_nlvr2} employs an additional step to manually remove non-interesting images, thus ensuring the content quality.
 
 \paragraph{Image subset construction.} \label{sec:dset_imgset}
 The nature of our task requires collections of negative images with high visual similarity, as otherwise, it would be trivial to discriminate between the reference and target image.
 Thus, prior to forming reference-target image pairs, we construct multiple subsets of six images that are semantically and visually similar, denoted as $\mathcal{S}=\left\lbrace I_1,\ldots,I_6 \right\rbrace$, shown in \figref{fig:dset_collection_0}(a).
 
 Here, to construct a subset, we randomly pick one image from the large corpus $I_1 \in \D$. We then sort the remaining images in $\D$ by their cosine similarity to $I_1$ using ResNet152~\cite{he2015deep} image feature vectors pre-trained on ImageNet~\cite{krizhevsky2012imagenet}. Denote by $\kappa_{i}$ the cosine similarity for image $I_i$. We then pick five additional images to produce a similar yet diverse subset, as follows: First, we filter out images with $\kappa_{i} \geq 0.94$ to avoid near-identical images to $I_1$. Then for the next top-20 ranked images, we greedily add each image in turn, skipping an image if its cosine similarity is within 0.002 of the last image added. If a subset of size six cannot be created, then the entire set is discarded.
 
Once constructed we further filter the collection subsets to avoid heavy overlap. We obtain in total 52,732 subsets from NLVR$^2$, from which we randomly choose 4,351 for the construction of \dstname.
 
 \paragraph{Image pairing.}
 Within each constructed image subset $\mathcal{S}$, we draw nine pairs of images, as shown in \figref{fig:dset_collection_0}(b).
 We choose these pairs to have (1) consecutive modifications that will allow future training of a dialogue systems; and (2) multiple outcomes from the same reference image.

 \begin{table*}[t!]
   \centering \scalebox{0.70}{
     \centering
     \renewcommand{\arraystretch}{1.05} 
 
     \begin{tabular}{ll cc l} 
       \toprule
       \multicolumn{1}{l}{\multirow{2}{*}{}} & \multicolumn{1}{l}{\multirow{2}{*}{Semantic aspect}} & \multicolumn{2}{c}{Coverage (\%)}         & \multirow{2}{*}{Example (boldface added here for emphasis)}                                                                    \\ 
       & & \dstname & \multicolumn{1}{c}{Fashion-IQ} &                                                                                             \\ 
       \midrule
       1 & Cardinality                                   & 29.3 & --                              & Only \textbf{one} of the boars and the ground is browner.                                   \\
       2 & Addition                                      & 15.2 & 15.7                          & \textbf{Add} human feet and a collar.                                                       \\
       3 & Negation                                      & 11.9 & ~4.0$^\dagger$                  & \textbf{Remove} the chair, make the dog sit in an open box.                                 \\ 
       \midrule
       4 & Direct Addressing                             & 57.4 & 49.0$^\dagger$                   & Show some lemons with a glass of lemonade.                                                  \\
       5 & Compare \& Change                               & 31.7 & ~3.0                           & \textbf{Same computer but} different finish and black background.                                  \\
       6 & Comparative Statement                                  & 51.7 & 32.0$^\dagger$                   & A \textbf{bigger} pub with \textbf{more} people on it.                                      \\
       7 & Statement with Conjunction                                & 43.7 & 19.0$^\dagger$                   & Remove all but one bird \textbf{and} have it facing right \textbf{and} putting food in its mouth.  \\ 
       \midrule
       8 & Spatial Relations \& Background               & 61.4 & --                              & Change the sky to blue color.                                                               \\
       9 & Viewpoint                                     & 12.7 & --                              & Focus widely on all available cookies package.                                              \\
       \specialrule{\heavyrulewidth}{2pt}{0pt}
       \rowcolor{Gray}
       & \textit{Avg. Sentence length (words)} & 11.3 & 5.3                              &                                               \\
       \bottomrule
       \end{tabular}} \\[3pt]
       \centering\footnotesize
       \begin{minipage}{0.98\linewidth}
         \centering
         {
           \frame{\includegraphics[height=4.90ex]{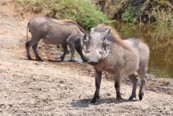}}~
           \frame{\includegraphics[height=4.90ex]{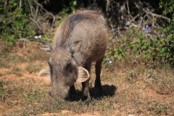}}~
           \frame{\includegraphics[height=4.90ex]{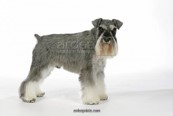}}~
           \frame{\includegraphics[height=4.90ex]{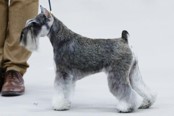}}~
           \frame{\includegraphics[height=4.90ex]{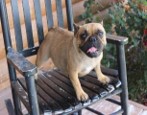}}~
           \frame{\includegraphics[height=4.90ex]{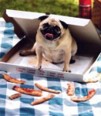}}~
           \frame{\includegraphics[height=4.90ex]{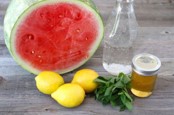}}~
           \frame{\includegraphics[height=4.90ex]{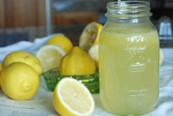}}~
           \frame{\includegraphics[height=4.90ex]{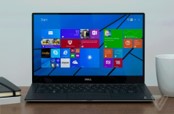}}~
           \frame{\includegraphics[height=4.90ex]{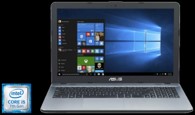}}~
           \frame{\includegraphics[height=4.90ex]{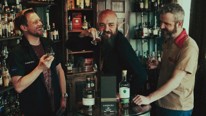}}~
           \frame{\includegraphics[height=4.90ex]{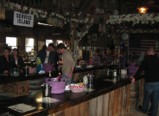}}~
           \frame{\includegraphics[height=4.90ex]{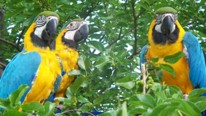}}~
           \frame{\includegraphics[height=4.90ex]{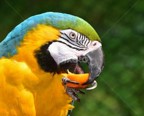}}~
           \frame{\includegraphics[height=4.90ex]{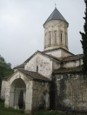}}~
           \frame{\includegraphics[height=4.90ex]{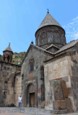}}~
           \frame{\includegraphics[height=4.90ex]{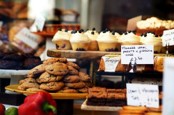}}~
           \frame{\includegraphics[height=4.90ex]{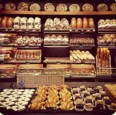}}~
         }\\
         \raggedright
         \begin{footnotesize}
           \textbf{1 \hspace{53pt}
                   2 \hspace{51pt}
                   3 \hspace{37pt}
                   4 \hspace{51pt}
                   5 \hspace{55pt}
                   6 \hspace{55pt}
                   7 \hspace{50pt}
                   8 \hspace{26pt}
                   9
           }
         \end{footnotesize}
       \end{minipage}
   \caption{Analysis of semantic aspects covered by the annotations in \dstname and in Fashion-IQ~\protect{\cite{fashioniq}}. We also show average sentence length (nb. words). $\dagger$ Numbers from~\protect{\cite{fashioniq}}. Image pair for each example is shown below with row number (left-right: reference-target).}
   \label{tab:dst_linguistic}
 \end{table*}

 \paragraph{Annotations.}\label{sec:dset_annotations}
 We collect a modification sentence for each pair of reference-target images using Amazon Mechanical Turk (AMT). To ensure that no false-negatives exist within the same image subset from which we draw the pair, as illustrated in \figref{fig:dset_collection_0}(c), we show AMT workers the remaining images from the subset and specifically ask them to write sentences that can \textit{only} lead to the true target image.
 
 AMT workers were instructed to avoid subjective descriptions, text mentions, plain side-by-side comparisons, or simple descriptions that only address the target images.

 Following the collection of the modification sentences for each pair, we additionally collect some auxiliary annotations that more explicitly address the ambiguities associated with implicit human-agreements. 
While we believe that these auxiliary annotations will be useful for future work, we do not make use them in our current work\footnote{See supp. mat. and our project website for details on auxiliary annotations.}.

 \paragraph{Data splits.}
 Following convention, we randomly assign 80\% of the data for training, 10\% for validation and 10\% for test. 
 Detailed statistics are shown in \tabref{tab:dst_stats}.

 \begin{table}[!ht]
  \centering \scalebox{0.70}{
    \begin{tabular}{l rrrr} 
      \toprule
      & Nb. image subsets & Nb. pairs & Nb. pairs per subset & Nb. images \\
      \midrule
      Train     & 3,345   & 28,225   & 7.54   & 16,939     \\
      Val.      & 503     & 4,184    & 8.32   & 2,297      \\
      Test      & 503     & 4,148    & 8.25   & 2,316      \\ 
      \midrule
      Total     & 4,351   & 36,554   & 8.40   & 21,552     \\
      \bottomrule
    \end{tabular}}
    \caption{Statistics of CIRR. Each reference-target image pair is associated with one annotation.}
    \label{tab:dst_stats}
\end{table}

 \subsection{Analysis on \dstname}
  
We follow Suhr \etal~\cite{Suhr_2019_nlvr2} and analyze coverage of various semantic concepts by keywords and sentence patterns (see \tabref{tab:dst_linguistic}). Here, we show comparisons with Fashion-IQ~\cite{fashioniq}, the most popular, comparable human-labeled dataset.
We observe a greater diversity and average length in the sentences in CIRR, indicating broad coverage and linguistic diversity.
Over 40\% of the annotations are compositional, which indicates an appreciable level of complexity of the sentences.
Interestingly, our annotations should also encourage models to attend to both the reference and target images by implicitly (rows 1--4) or explicitly (rows 5--6) referring to the visual contents of \emph{both} images.

\section{Experiments}

\begin{table*}[!ht]
  \centering \scalebox{0.70}{
  \begin{tabular}{lllrrrrrrrr} 
  \toprule
  \multicolumn{2}{c}{} & \multicolumn{1}{c}{}        & \multicolumn{4}{c}{Recall$@K$ }               & \multicolumn{3}{c}{Recall$_{\text{Subset}}@K$ }   & \multirow{2}{*}{(R$@5$ $+$ R$_{\text{Subset}}@1$)$/2$}       \\
  \cmidrule(lr){4-7}
  \cmidrule(lr){8-10}
  \multicolumn{2}{l}{} & \multicolumn{1}{l}{Methods} & $K=1$           & $K=5$           & $K=10$ & $K=50$          & $K=1$           & $K=2$           & $K=3$            &  \\ 
  \midrule
  \multirow{8}{*}{\rotatebox{90}{\sc Baselines}}
  & 1 & Random~(theoretical)              & 0.02             & 0.12            & 0.24 & 1.20            & 20.00           & 40.00           & 60.00         & 10.06  \\
  & 2 & Random~(init. ResNet)              & 7.18             & 25.74          & 36.91 & 66.68           & 20.84           & 41.02           & 61.65        & 23.29   \\ 
  \cmidrule{2-11}
  & 3 & Image-only                  & 13.73            & {\color{blue}\textbf{48.46}}                         & {\color{blue}\textbf{65.81}} & 89.94   & 20.93           & 42.15           & 63.26         & 34.70  \\
  & 4 & Text-only                   & 3.90             & 13.17           & 20.43 & 49.16           & \textbf{39.69}   & 62.23   & 78.52   & 26.43 \\
  & 5 & Random Image$+$Text                   & 2.99             & 11.91           & 19.85 & 46.97           & 39.41   & {\color{blue}\textbf{62.33}}   & {\color{blue}\textbf{78.71}}   & 25.66\\
  & 6 & Image+Text Concatenation                      & 12.44                     & 40.24                     & 57.52 & 87.29           & 23.74           & 45.12           & 65.50      & 31.99\\ 
  \cmidrule{2-11}
  & 7 & Human Performance$^\dagger$                      & --                     & --                     & --  & --           & \uline{86.09}           & --           & --        &  --  \\ 
  \midrule[\heavyrulewidth]
  \multirow{7}{*}{\rotatebox{90}{\sc SoTA}}
  & 8 & TIRG~\cite{Vo_2019_tirg}             & 14.61  & 48.37  & 64.08  & {\color{blue}\textbf{90.03}} &  22.67  & 44.97  & 65.14  & 35.52 \\ 
  & 9 & TIRG$+$LastConv~\cite{Vo_2019_tirg}             & 11.04  & 35.68  & 51.27  & 83.29 &  23.82  & 45.65  & 64.55  & 29.75 \\ 
  \cmidrule{2-11}
  & 10 & MAAF~\cite{dodds2020modality_maaf}         & 10.31     & 33.03     & 48.30    & 80.06 &  21.05    & 41.81    & 61.60  & 27.04 \\
  & 11 & MAAF$+$BERT~\cite{dodds2020modality_maaf}  & 10.12     & 33.10     & 48.01 & 80.57    & 22.04    & 42.41    & 62.14   & 27.57 \\
  & 12 & MAAF$-$IT~\cite{dodds2020modality_maaf}    & 9.90      & 32.86     & 48.83 & 80.27    & 21.17    & 42.04    & 60.91   & 27.02 \\
  & 13 & MAAF$-$RP~\cite{dodds2020modality_maaf}    & 10.22     & 33.32     & 48.68 & 81.84    & 21.41    & 42.17    & 61.60   & 27.37 \\
  \midrule
  & 14 & Ours (no init.)    & {\color{blue}\textbf{15.18}}  & 43.36  & 60.48  & 87.64  & 33.81   & 56.99  & 75.40  & {\color{blue}\textbf{38.59}} \\ 
  & 15 & Ours (init. OSCAR) & \textbf{19.55}  & \textbf{52.55}  & \textbf{68.39}  & \textbf{92.38} &  {\color{blue}\textbf{39.20}}  & \textbf{63.03}  & \textbf{79.49}  & \textbf{45.88} \\ 

  \bottomrule
  \end{tabular}}
  \caption{Retrieval performance on CIRR. Best (resp. second-best) numbers are in bold-black (resp. blue). $\dagger$ See supplementary material on our collection details of human performance. We additionally report the average score over R$@5$ and R$_{\text{Subset}}@1$, which better reveals the overall performance of models (discussed in \secref{sec:baseline_comparison}). Note that R$@5$ accounts for possible false-negatives in the entire image corpus. Since R$_{\text{Subset}}$ is not affected by such issues (\secref{sec:extended_metric}), we consider R$_{\text{Subset}}@1$ to better illustrate the fine-grained reasoning ability of methods.}
  \label{tab:baseline_0}
\end{table*}

\begin{table*}[!ht]
  \centering \scalebox{0.7}{
  \begin{tabular}{llrrrrrrrrr} 
  \toprule
  \multicolumn{1}{c}{} & \multicolumn{1}{c}{} & \multicolumn{2}{c}{Dress} & \multicolumn{2}{c}{Shirt} &\multicolumn{2}{c}{Toptee} &\multicolumn{2}{c}{Avg} & \multirow{2}{*}{(R$@10$ + R$@50$)$/2$} \\
  \cmidrule(lr){3-4}
  \cmidrule(lr){5-6}
  \cmidrule(lr){7-8}
  \cmidrule(lr){9-10}
  \multicolumn{1}{l}{} & \multicolumn{1}{l}{Methods} & R$@10$ & R$@50$ & R$@10$ & R$@50$ & R$@10$ & R$@50$ & R$@10$ & R$@50$ &  \\ 
  \midrule
  1 & Image-only                & 4.20    & 13.29          & 4.51 & 14.47           & 4.13  & 14.30    & 4.28 & 14.20 & 9.15            \\ 
  2 & Image+Text Concatenation  & 10.52    & 28.98          & 13.44 & 34.60           & 11.36  & 30.42    & 11.77 & 31.33 & 21.55            \\
  3 & TIRG~\cite{Vo_2019_tirg}  & 8.10    & 23.27          & 11.06 & 28.08           & 7.71  & 23.44    & 8.96 & 24.93 & 16.95            \\
  4 & TIRG+Side Information~\cite{fashioniq}    &11.24    & 32.39          & 13.73 & 37.03           & 13.52  & 34.73    & 12.82 & 34.72 & 23.77    \\
  \specialrule{\lightrulewidth}{2pt}{0pt}
  \rowcolor{Gray}
  5 & MRN~\cite{MRN}    &12.32    & 32.18          & 15.88 & 34.33           & 18.11  & 36.33    & 15.44 & 34.28 & 24.86    \\
  \rowcolor{Gray}
  6 & FiLM~\cite{perez2017film}    &14.23    & 33.34          & 15.04 & 34.09           & 17.30  & 37.68    & 15.52 & 35.04 & 25.28    \\
  \rowcolor{Gray}
  7 & TIRG~\cite{Vo_2019_tirg}    &14.87    & 34.66          & 18.26 & 37.89           & 19.08  & 39.62    & 17.40 & 37.39 & 27.40    \\
  \rowcolor{Gray}
  8 & Relationship~\cite{santoro2017simple}    &15.44    & 38.08          & 18.33 & 38.63     & 21.10  & 44.77    & 18.29 & 40.49 & 29.39    \\
  \specialrule{\lightrulewidth}{0pt}{0pt}

  \rowcolor{Gray}
  9 & VAL (init. GloVe)~\cite{chen2020image_val}       &22.53    & 44.00          & 22.38 & 44.15     & 27.53  & 51.68    & 24.15 & 46.61 & 35.40    \\
  10 & MAAF~\cite{dodds2020modality_maaf} &\textbf{23.8}\phantom{0}    & \textbf{48.6}\phantom{0}          & \textbf{21.3}\phantom{0} & \textbf{44.2}\phantom{0}     & \textbf{27.9}\phantom{0}  & \textbf{53.6}\phantom{0}    & \textbf{24.3}\phantom{0} & \textbf{48.8}\phantom{0} & \textbf{36.6}\phantom{0}    \\

  \midrule
  13 & Ours (no init.)    &14.38   & 34.66     & 13.64 & 33.56     & 16.44  & 38.34   & 14.82 & 35.52 & 25.17  \\ 
  14 & Ours (init. OSCAR) &17.45   & 40.41     & 17.53 & 38.81     & 21.64  & 45.38   & 18.87 & 41.53 & 30.20   \\ 

  \bottomrule
  \end{tabular}}
  \caption{Retrieval performance on Fashion-IQ, we follow~\cite{fashioniq} to report average scores of R$@10$ and 50. Best numbers for SoTA models are in bold-black. Rows 1-4 reported by~\cite{fashioniq}, rows 5-9 (shaded) reported by~\cite{chen2020image_val}. Rows 9-10 are SoTA methods developed for composed image retrieval, where we report the originally published numbers of their best configurations. Note that we see multiple scores reported for TIRG on Fashion-IQ, here we only show the published results from the above two sources. Additional non peer-reviewed methods that involve ensembles of models or data augmentation are not included.}
  \label{tab:baseline_1}
\end{table*}

\paragraph{Datasets.}
To demonstrate the model's ability in untilizing pre-trained V\&L knowledge, as well as its generalizability to images of different domains, we evaluate our proposed model against baselines and state-of-the-art (SoTA) methods on two datasets, including \textbf{(1)} \dstname, our proposed dataset on open-domain composed image retrieval, and \textbf{(2)} Fashion-IQ~\cite{fashioniq}, which contains images of fashion products among three subtypes (\texttt{Dress}, \texttt{Shirt}, \texttt{Toptee}) with human-generated annotations. We do not evaluate on other datasets discussed in \secref{sec:related_work}, as they either contain synthetic image/annotation or are domain-wise similar to Fashion-IQ (\eg, Fashion200k~\cite{han2017automatic_fashion200k}).

\paragraph{Compared methods.}
For \dstname, we evaluate the following methods using publicly available  implementations\footnote{\url{https://github.com/google/tirg}, \url{https://github.com/yahoo/maaf}}:
\begin{itemize}
  \denselist
  \item TIRG~\cite{Vo_2019_tirg} is an image-text composition model for composed image retrieval, which has proven to be effective on multiple datasets~\cite{fashioniq,han2017automatic_fashion200k,Vo_2019_tirg,Isola2015DiscoveringSA_mitstates}. The method uses a gating and residual design to encourage the learning of cross-modal features. Two setups for TIRG are available based on whether to inject text features at the last FC-layer (\textbf{default}), or the last convolution layer (\textbf{LastConv}). We test both setups.
  \item MAAF~\cite{dodds2020modality_maaf} is specifically designed for composed image retrieval with state-of-the-art performance. By default, it treats the convolutional spatial image features and the learned text embeddings (randomly initialized with LSTM~\cite{hochreiter1997long}) as modality-agnostic tokens, which are passed to a Transformer~\cite{vaswani2017attention_transformer}. We evaluate three design choices that were originally reported with comparable results: {\bf (+BERT)}~pretrained context-aware word representations using BERT~\cite{devlin2018bert_bert}, {\bf  (-IT)}~removing the output of text tokens in the last pooling layer, {\bf (-RP)}~substituting the final resolution-wise pooling with average pooling.
\end{itemize}

For comparison, we also evaluate the following baselines, implemented by \citet{Vo_2019_tirg}:
\begin{itemize}
  \denselist
  \item Random~(theoretical): theoretical random guess.
  \item Random~(init. ResNet): pretrained ImageNet~\cite{krizhevsky2012imagenet} features, but random weights for others parameters.
  \item Image and text-only: substituting the combined image-text feature with the reference image or text feature.
  \item Random image with text: randomly sampling images to pair with text during training and validation.
  \item Concatenation: replacing the image-text composition layer with a simple concatenation of features followed by a 2-layer perceptron with ReLU.
\end{itemize}

For Fashion-IQ, we additionally include published results from the following methods:
\begin{itemize}
  \denselist
  \item MRN~\cite{MRN} uses stacked blocks of element-wise products with residual learning to embed V\&L jointly.
  \item FiLM~\cite{perez2017film} modulates the image feature map conditioned on text features after the layers of CNN.
  \item Relationship~\cite{santoro2017simple} learns the joint embeddings through relationship features constructed by concatenating the image and text features followed by FC-layers.
  \item VAL~\cite{chen2020image_val} is specially designed for composed image retrieval, which adopts the Transformer to compose multi-level V\&L joint representations. For images with text descriptions as side information, an additional visual-semantic loss is applied to align visual features and the corresponding text features.
\end{itemize}

\paragraph{Metric.}\label{sec:extended_metric}
We follow previous work to report retrieval performance in Recall within top-$K$ (Recall$@K$).
For \dstname, we additionally report Recall$_{\text{subset}}$, which is an extension to the standard (global) Recall, made possible by the unique design of our dataset.

As discussed, our input queries $q = \langle I_{\text{R}}, t\rangle$ and target images $I_{\text{T}}$ in our dataset are constructed such that both $I_{\text{R}}$ and $I_{\text{T}}$ are sampled from the same image set $\mathcal{S}$ (\secref{sec:dset_imgset}).
We formulate Recall$_{\text{subset}}$ task by ranking images in $\mathcal{S} \setminus \{ I_{\text{R}} \}$ according to model score. We define Recall$_{\text{subset}}@K$ as the proportion of (test) examples where the ground-truth target image $I_{\text{T}}$ is ranked within the top-$K$ image in its subset.

Conceptually, Recall$_{\text{subset}}$ can be viewed as Recall while only considering images within the same subset as the pair. 
The benefits are twofold: First, Recall$_{\text{subset}}$ is not affected by false-negative samples, thanks to our careful design in data collection procedures. Second, with a selected batch of negative samples with high visual similarities, Recall$_{\text{subset}}$ can facilitate analysis on the reasoning ability of the methods for capturing fine-grained image-text modifications.

\paragraph{Implementation details.}\label{sec:implementation_detail}
All experiments are conducted on a single NVIDIA RTX3090 with PyTorch. SoTA models use the default configurations proposed by their authors. 
See supp. mat. and our project website for more details on baseline training.
For our proposed model, we use ResNet152 for image feature extraction. The model is optimized with AdamW~\cite{loshchilov2018decoupled} with an initial learning rate of $10^{-5}$. We set a linearly decreasing schedule without warm-up. The batch size is set to 32 and the network is trained for 300 epochs. Other settings are kept as default by OSCAR.

\subsection{Results}

\begin{figure*}[t!]
  \begin{center}
    \includegraphics[trim={0 1pt 0 0pt},clip, width=0.995\linewidth]{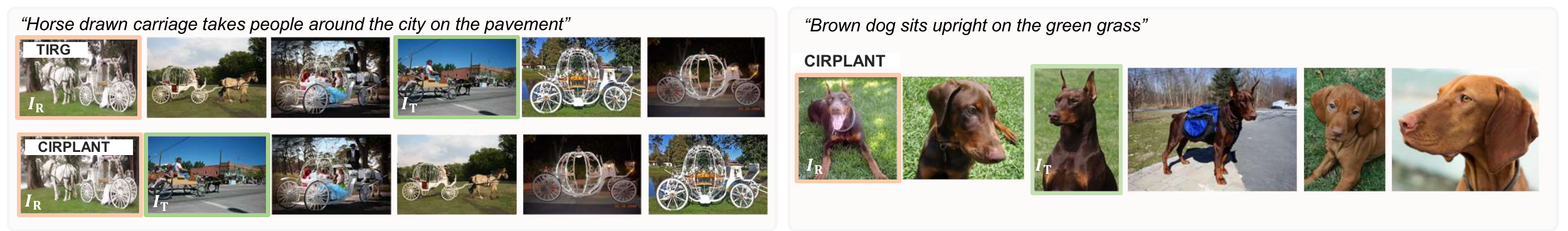}
  \end{center}
  \caption{Qualitative results of image retrieval on \dstname, red/green boxes: reference/target images. Predictions are ranked from left to right. We show the ranked images \uline{within subsets}, see \secref{sec:extended_metric} for details on metric. (Left) We compare the retrieval on the same query for TIRG and \modelname. (Right) We demonstrate the implicit ambiguities within the dataset (in this case, the difficulty in selecting the \textit{most suitable} candidate by preserving the breed of the dog across the images, which requires identifying subtle characteristics-- \eg pointy ears).}
  \label{fig:quali-0}
  \vspace{-1.5em}
\end{figure*}

\paragraph{Baseline comparison on \dstname.}\label{sec:baseline_comparison}
\tabref{tab:baseline_0} (rows 1-13) compares the retrieval performance of baseline and SoTA methods for both Recall and Recall$_{\text{Subset}}@K$ on CIRR. 

For global Recall, we notice that TIRG performs similar to the Image-only baseline, suggesting that its multi-modal composition layers often fail to extract information from the text. Instead, it relies primarily on visual content. We conjecture that CIRR focuses more on the fine-grained changes that are harder to capture and associate across modalities, therefore, requires stronger image-text composition layers. In addition, we note that MAAF (rows 10-13) does not generalize well to our dataset, even though it outperforms TIRG and other methods on existing ones~\cite{dodds2020modality_maaf}. We believe the choice of forming image tokens by spatial feature maps does not generalize to our dataset where the modification concepts are more diverse and at multiple levels. Meanwhile, adding the contextual-aware BERT pretrained weights yields little effects, suggesting a plain initialization of word embeddings, though contains validated pre-trained language information, may not help the composition layers.

The Recall$_{\text{Subset}}$ results tell a similar story. Here the performance of all SoTA models is close to the theoretical random guess, indicating that current models fail to capture fine-grained modifications between similar images.
Interestingly, we discover that the Text-only and Random-Image$+$Text baselines (rows 4,5) outperform SoTA models significantly. We believe this is because the modification sentences usually contain descriptions of visual content that is unique to the target image once limited to the smaller retrieval set (\eg, ``add a leash to the dog'' where only the target image contains the leash).
However, as demonstrated by the low Recall performance, such descriptions are not detailed enough to single out the target image in the entire image corpus.
This scenario further demonstrates Recall$_{\text{Subset}}$ reveals behaviors of models on different aspects, and can be used for more detailed analysis. 

In short, the relatively low retrieval performance suggests that our dataset poses a challenge to existing methods developed and tested on narrow-domain datasets.

\paragraph{Performance of \modelname on \dstname.}
Results in \tabref{tab:baseline_0} (rows 14,15) compares our proposed model with SoTA methods on \dstname.
We notice that on \dstname, \modelname with no initialization (row 14) performs similarly as TIRG on Recall, while surpassing all other SoTA methods. This validates our design choice of using non-regional image features for composing image and text through the transformer architecture. Meanwhile, on Recall$_{\text{Subset}}$ our model, even without initialization, yields much higher scores than others, suggesting transformers are better in capturing more fine-grained visiolinguistic cues when composing image and text features. Comparing with SoTA methods that use LSTMs for generating a single language embedding of the entire sentence, we believe that the key difference lies within the fact that transformers accept word tokens as input, which can later be attended individually.
Our model outperforms all other methods with OSCAR initialization (row 15) by a significant margin, demonstrating the benefit of VLP knowledge on open-domain images.

\paragraph{Performance of \modelname on Fashion-IQ.}
\tabref{tab:baseline_1} compares the performance of our model with SoTA methods. We notice that our model with OSCAR initialization (row 14) outperforms most methods, including 
generic multimodal learning methods and TIRG. 
This strengthens the benefits of using transformer architecture that leverages VLP models.
Additionally, we note that even on Fashion-IQ, our model still benefits greatly from OSCAR pre-trained initialization (rows 13,14). Given that the images in Fashion-IQ differ greatly from the data used for OSCAR pre-training~\cite{oscar}, we believe this further demonstrates that the pre-trained model can transfer the learned V\&L knowledge and adapt to various contexts.

We note that two recent SoTA methods for composed image retrieval (VAL and MAAF, rows 9,10) perform better than our model. Despite the visible improvements brought by OSCAR initialization, we hypothesize that our model is still underperformed by the apparent domain shift in images, as the VLP model is pre-trained on generic ImageNet-type data.
Meanwhile, the low generalizability of MAAF on \dstname (\tabref{tab:baseline_0} rows 10-13) hints the possibility that current SoTA methods developed and tested on existing datasets may have been overly adapted to domain-specific images of low complexity. Hence, additional open-domain datasets, such as \dstname, can be beneficial in future research.

\subsection{Qualitative Results}\label{sec:quali}
\figref{fig:quali-0} (left) demonstrates the retrieval rankings within the image subset (see \secref{sec:extended_metric}) on the same query for TIRG and \modelname. Specifically, we show the effectiveness of pre-training in \modelname when encountering visiolinguistic concepts (\ie, \textit{pavement}) that occur less frequently in the training data. Additionally, \modelname better captures fine-grained cues within language (\eg, \textit{takes people around}, which implies \textit{must have people in the back of the carriage}), thanks to the transformer architecture that accepts, and attends to individual word tokens.

We show one failure case of \modelname on \dstname in \figref{fig:quali-0} (right). Note the implicit requirement of \textit{preserving same breed of dog} across the reference and target image. This requires models to identify the fine-grained visiolinguistic cues (\ie, pointy ears in this sample) and retrieve the most suitable image, bringing more challenge to the task. 

\section{Conclusion}
This work expands the task of composed image retrieval into more complex, open-domain images. We collect the \dstname dataset, which addresses shortcomings of existing datasets by placing more emphasis on distinguishing open-domain visually similar images.
Our publicly available dataset is designed to facilitate future studies on subtle reasoning over visiolinguistic concepts, as well as iterative retrieval with dialogue.
We also introduce \modelname, a transformer-based model that leverages V\&L pre-training to compose image and text features. We validate \modelname on both \dstname and the existing fashion dataset, demonstrating the generalizability of our design and the effectiveness of V\&L pre-training.
Collectively, we hope to inspire future work on composed image retrieval on a broader scope, yet fine-grained level.

{\small
\bibliographystyle{abbrvnat}
\bibliography{egbib}
}

\clearpage
\appendix
\section*{Supplementary Material}
\section{Implementation Details}

\paragraph{Existing methods on \dstname.}
As discussed in \secref{sec:implementation_detail}, we adopt the default configurations for state-of-the-art (SoTA) methods when testing on our proposed dataset \dstname.

Specifically, for TIRG~\cite{Vo_2019_tirg} and its corresponding baselines (incl. Random, Image/text-only, Random Image$+$Text and Concatenation), we use ResNet18 pretrained on ImageNet~\cite{krizhevsky2012imagenet} as the image encoder, and a randomly initialized LSTM as the text encoder. We note that the above methods do not benefit from more complex ResNet features (\eg, ResNet152) or word embedding initializations on \dstname.
We train the models using soft-triplet based loss~\cite{Vo_2019_tirg}, as we discover that the batch-based classification loss introduces serious overfitting for TIRG on \dstname. 
For MAAF, following~\citet{dodds2020modality_maaf}, we use a pretrained ResNet50 along with an LSTM, while training the model with batch-based classification loss. 

Note that the implementations of MAAF and TIRG share the same codebase. Hence, all methods above use a hidden size of 512, and models are optimized with vanilla stochastic gradient descent (SGD) as in Vo \etal~\cite{Vo_2019_tirg}.

\paragraph{\modelname on Fashion-IQ.}
We do not perform hyper-parameter tuning for our model on Fashion-IQ (\ie, the setup is kept the same as on \dstname, see \secref{sec:implementation_detail} for details). 
Additionally, since the three subtypes in Fashion-IQ distinct greatly from each other, we sample each minibatch from a single subtype during training, as in~\citet{dodds2020modality_maaf}.

\section{Additional Metrics}
See \tabref{tab:reb_map_0} on performance in mAP$@K$, where the comparisons are similar to Recall (\tabref{tab:baseline_0}).
Note that since our task has only one true-positive for each query, Precision$@K$ and Recall$@K$ are the same (hence P$@K$ not shown).

\begin{table}[!ht]
  \centering 
    \scalebox{0.58}{
    \begin{tabular}{lllrrrrrrr}
      \toprule
      \multicolumn{2}{c}{} & \multicolumn{1}{c}{}        & \multicolumn{4}{c}{mAP$@K$ }           & \multicolumn{3}{c}{mAP$_{\text{Subset}}@K$ }                                                                                                                                                                                           \\
      \cmidrule(lr){4-7}
      \cmidrule(lr){8-10}
      \multicolumn{2}{l}{} & \multicolumn{1}{l}{Methods}                                             & $K=1$                                     & $K=5$                                           & $K=10$                       & $K=50$                       & $K=1$                        & $K=2$                        & $K=3$                                                       \\
      \midrule
                           & 8                           & TIRG~\cite{Vo_2019_tirg}                  & 14.61                                           & 25.05 & 27.25                        & 28.63 & 22.67                        & 33.25                        & 39.92                        \\
                           & 10                           & MAAF~\cite{dodds2020modality_maaf}        & 10.31                                           & 15.77                        & 17.72                        & 19.43                        & 21.05                        & 29.84                        & 36.80                        \\
      \midrule
                           & 14                          & Ours (no init.)                        & 15.18                               &  25.06             &  27.19              & 28.65               &  33.81              &  45.50            &  51.60              \\
                           & 15                          & \textbf{Ours (init.)}                        & \textbf{19.55}                                  & \textbf{30.40}               & \textbf{32.55}               & \textbf{33.85}               & \textbf{39.20}               & \textbf{50.68}               & \textbf{56.28}               \\

      \bottomrule
    \end{tabular}}
  \caption{mAP scores for SoTA methods (and ours) on CIRR. See \tabref{tab:baseline_0} (corresp. row numbers) for comparison with Recall.}
  \label{tab:reb_map_0}
\end{table}
\par\vspace{-1.0em}\par

\section{Auxiliary Annotations in \dstname} \label{sec:aux_anno}
As discussed in \secref{sec:dset_annotations}, following the collection of the modification sentences (main annotation), we additionally collect auxiliary annotations for each image pair. 
The auxiliary annotations are meant to provide explicit training signals that address the ambiguities caused by implicit human-agreements. 
Although we do not use such annotations in this work, we believe that they can benefit future work for clarifying and interpreting such ambiguities.

\paragraph{Collection.}
For each pair of reference-target image, we collect the answers to the following four questions from Amazon Mechanical Turk (AMT) workers, which tangibly address the implicit ambiguities mentioned above:
\begin{itemize}
  \denselist
  \item[\bf Q1] What characteristics of the objects are preserved across images?
  \item[\bf Q2] What objects were changed, but not relevant to the modifying sentence?
  \item[\bf Q3] Is there any change in camera angle/focus/viewpoint?
  \item[\bf Q4] Is there any change in the background/lighting across the images?
\end{itemize}

We provide the AMT workers with the reference-target image pair along with the collected modification sentence (main annotation).
For each question, workers can choose to answer with a sentence or mark as not applicable (\eg, nothing worth mentioning or already covered by the main annotation).
Statistics are shown in \tabref{tab:dst_stats_ext}. Collection interface is shown in \figref{fig:amt_interface} (bottom), see examples in \figref{fig:cirr_examples}.

\begin{table}[!ht]
  \centering \scalebox{0.65}{
    \begin{tabular}{l rrrrrrrr} 
      \toprule
      & \multicolumn{1}{r}{\multirow{2}{*}{\begin{tabular}[c]{@{}r@{}}Nb. image\\subsets\end{tabular}}} & \multicolumn{1}{r}{\multirow{2}{*}{\begin{tabular}[c]{@{}r@{}}Nb.\\pairs\end{tabular}}} & \multicolumn{1}{r}{\multirow{2}{*}{\begin{tabular}[c]{@{}r@{}}Nb. pairs\\per subset\end{tabular}}} & \multicolumn{1}{r}{\multirow{2}{*}{\begin{tabular}[c]{@{}r@{}}Nb.\\images\end{tabular}}} & \multicolumn{4}{c}{~~Pairs with auxiliary (\%)}  \\ 
      \cmidrule(lr){6-9}
      & & & & & Q1~~    & Q2~~    & Q3~~    & Q4~~ \\ 
      \midrule
      Train                & 3,345                                                                                        & 28,225                                                                                                    & 7.54                                                                                                & 16,939                                      & 66.72 & 68.09 & 48.06 & 58.45                                  \\
      Val.                 & 503                                                                                          & 4,184                                                                                                     & 8.32                                                                                                & 2,297                                       & 71.87 & 67.67 & 49.43 & 64.66                                  \\
      Test               & 503                                                                                          & 4,148                                                                                                     & 8.25                                                                                                & 2,316                                       & 69.62 & 69.01 & 46.44 & 63.00                                  \\ 
      \midrule
      Total                & 4,351                                                                                        & 36,554                                                                                                    & 8.40                                                                                                & 21,552                                      & 67.65 & 68.15 & 48.02 & 59.69                                  \\
      \bottomrule
    \end{tabular}}
    \caption{Statistics of CIRR \underline{with auxiliary annotations}. The visual contents and the (main) annotation determine whether a pair also has auxiliary annotations for Q1--4.}
    \label{tab:dst_stats_ext}
\end{table}

\section{Collection Details on \dstname}
We provide additional details about our data collection procedure (\secref{sec:dset_annotations}) including examples of each step (excl. the auxiliary annotations, which is discussed in \secref{sec:aux_anno}).

\paragraph{Image subsets.}\label{sec:sup_subset}
\shadowoffset{2pt}
\begin{figure*}[p]
  \centering\footnotesize
  \begin{minipage}{0.98\linewidth}
    \centering
    \setlength{\tabcolsep}{2pt}
    \begin{tabular}{|c|ccccccccc|}
      \multicolumn{10}{p{0.99\linewidth}}{} \\[3.5ex]
      \multicolumn{10}{p{0.99\linewidth}}{
        \textbf{(a)} Randomly pick an image as $I_\text{1}$ (leftmost), sort the remaining images in the large image corpus $\D$ by their cosine similarity to $I_\text{1}$ using ResNet features pre-trained on ImageNet, noted as $\kappa_i$ for $I_i$. Images are ranked from left to right.
      }\\
      \hline
      $I_\text{1}$&
      &      
      &      
      &      
      &      
      &      
      &      
      &
      &
      \\
      \frame{\includegraphics[height=8.0ex]{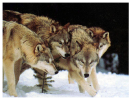}}& 
      \frame{\includegraphics[height=8.0ex]{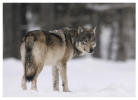}}&
      \frame{\includegraphics[height=8.0ex]{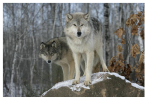}}&
      \frame{\includegraphics[height=8.0ex]{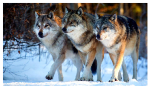}}&
      \frame{\includegraphics[height=8.0ex]{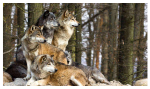}}&
      \frame{\includegraphics[height=8.0ex]{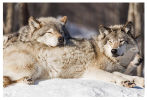}}&
      \frame{\includegraphics[height=8.0ex]{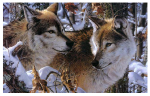}}&
      \frame{\includegraphics[height=8.0ex]{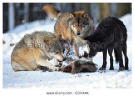}}&
      \frame{\includegraphics[height=8.0ex]{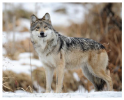}}&
      \frame{\includegraphics[height=8.0ex]{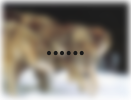}} \\
      $\kappa_i=1.0$&
      0.9981&      
      0.8691&      
      0.8663&      
      0.8603&      
      0.8490&      
      0.8488&
      0.8456&
      0.8435&
      ... \\
      \hline
      \multicolumn{10}{p{0.99\linewidth}}{} \\[0.5ex]
      \multicolumn{10}{p{0.99\linewidth}}{
        \textbf{(b)} Remove near-identical images with $\kappa_i\geq0.94$.
      }\\
      \hline
      $I_\text{1}$&
      \textbf{Removed}&      
      &      
      &      
      &      
      &      
      &      
      &
      &
      \\
      \frame{\includegraphics[height=8.0ex]{imgs_sup/exp0-0}}& 
      \frame{
        \begin{overpic}[height=8.0ex]{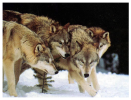}
          \put (60,33) {\textcolor{BrickRed}{\shadowtext{\huge\xmark}}}
        \end{overpic}
      }&  
      \frame{\includegraphics[height=8.0ex]{imgs_sup/exp0-3}}&
      \frame{\includegraphics[height=8.0ex]{imgs_sup/exp0-4}}&
      \frame{\includegraphics[height=8.0ex]{imgs_sup/exp0-5}}&
      \frame{\includegraphics[height=8.0ex]{imgs_sup/exp0-6}}&
      \frame{\includegraphics[height=8.0ex]{imgs_sup/exp0-7}}&
      \frame{\includegraphics[height=8.0ex]{imgs_sup/exp0-8}}&
      \frame{\includegraphics[height=8.0ex]{imgs_sup/exp0-9}}&
      \frame{\includegraphics[height=8.0ex]{imgs_sup/exp0-n}} \\
      $\kappa_i=1.0$&
      \textbf{0.9981}&      
      0.8691&      
      0.8663&      
      0.8603&      
      0.8490&      
      0.8488&
      0.8456&
      0.8435&
      ... \\
      \hline
    \multicolumn{10}{p{0.99\linewidth}}{} \\[0.5ex]
      \multicolumn{10}{p{0.99\linewidth}}{
        \textbf{(c)} Select the next top-20 ranked images (not fully shown below).
      } \\
      \hline
      $I_\text{1}$    
      & \multicolumn{2}{c}{$\Leftarrow$ Shifted} 
      &
      &      
      &      
      &      
      &
      &
      &
      $\Rightarrow$ top-20\\
      \frame{\includegraphics[height=8.0ex]{imgs_sup/exp0-0}}& 
      \frame{\includegraphics[height=8.0ex]{imgs_sup/exp0-3}}&
      \frame{\includegraphics[height=8.0ex]{imgs_sup/exp0-4}}&
      \frame{\includegraphics[height=8.0ex]{imgs_sup/exp0-5}}&
      \frame{\includegraphics[height=8.0ex]{imgs_sup/exp0-6}}&
      \frame{\includegraphics[height=8.0ex]{imgs_sup/exp0-7}}&
      \frame{\includegraphics[height=8.0ex]{imgs_sup/exp0-8}}&
      \frame{\includegraphics[height=8.0ex]{imgs_sup/exp0-9}}&
      \frame{\includegraphics[height=8.0ex]{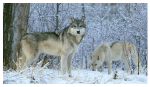}}&
      \frame{\includegraphics[height=8.0ex]{imgs_sup/exp0-n}} \\
      $\kappa_i=1.0$&
      0.8691&      
      0.8663&      
      0.8603&      
      0.8490&      
      0.8488&
      0.8456&
      0.8435&
      0.8421&
      ... \\
      \hline
    \multicolumn{10}{p{0.99\linewidth}}{} \\[1.0ex]
      \multicolumn{10}{p{0.99\linewidth}}{
        \textbf{(d)} Greedily add each image as ranked. Meanwhile, to ensure sufficient variations between images, skip an image if its $\kappa_i$ is within 0.002 of the last added image. We demonstrate the greedy process as below. In each step, curved arrow suggests a comparison of $\kappa_i$ and $\kappa_{i+1}$, added image is marked with a tick while skipped is crossed out.
      } \\
      \hline
      $I_\text{1}$&    
      $I_\text{2}$&      
      &      
      &      
      &      
      &      
      &
      &
      &
      \\
      \frame{
        \begin{overpic}[height=8.0ex]{imgs_sup/exp0-0}
          \put (55,33) {\textcolor{ForestGreen}{\shadowtext{\huge\cmark}}}
        \end{overpic}
      }& 
      \frame{
        \begin{overpic}[height=8.0ex]{imgs_sup/exp0-3}
          \put (55,30) {\textcolor{ForestGreen}{\shadowtext{\huge\cmark}}}
        \end{overpic}
      }&
      \frame{\includegraphics[height=8.0ex]{imgs_sup/exp0-4}}&
      \frame{\includegraphics[height=8.0ex]{imgs_sup/exp0-5}}&
      \frame{\includegraphics[height=8.0ex]{imgs_sup/exp0-6}}&
      \frame{\includegraphics[height=8.0ex]{imgs_sup/exp0-7}}&
      \frame{\includegraphics[height=8.0ex]{imgs_sup/exp0-8}}&
      \frame{\includegraphics[height=8.0ex]{imgs_sup/exp0-9}}&
      \frame{\includegraphics[height=8.0ex]{imgs_sup/exp0-10}}&
      \frame{\includegraphics[height=8.0ex]{imgs_sup/exp0-n}} \\
      $\kappa_i=1.0$&
      0.8691&      
      &      
      &      
      &      
      &
      &
      &
      &
      ... \\
      \hline
      \multicolumn{3}{c}{\includegraphics[height=2.5ex]{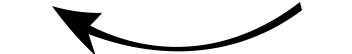}} ${\scriptstyle 1.0-0.8691>0.002}$&  
      &  
      &  
      &  
      &  
      &  
      & \multicolumn{1}{l}{} \\
      \hline
      $I_\text{1}$&    
      $I_\text{2}$&      
      $I_\text{3}$&      
      &      
      &      
      &      
      &
      &
      &
      \\
      \frame{
        \begin{overpic}[height=8.0ex]{imgs_sup/exp0-0}
          \put (55,33) {\textcolor{ForestGreen}{\shadowtext{\huge\cmark}}}
        \end{overpic}
      }& 
      \frame{
        \begin{overpic}[height=8.0ex]{imgs_sup/exp0-3}
          \put (55,30) {\textcolor{ForestGreen}{\shadowtext{\huge\cmark}}}
        \end{overpic}
      }&
      \frame{
        \begin{overpic}[height=8.0ex]{imgs_sup/exp0-4}
          \put (55,20) {\textcolor{ForestGreen}{\shadowtext{\huge\cmark}}}
        \end{overpic}
      }&
      \frame{\includegraphics[height=8.0ex]{imgs_sup/exp0-5}}&
      \frame{\includegraphics[height=8.0ex]{imgs_sup/exp0-6}}&
      \frame{\includegraphics[height=8.0ex]{imgs_sup/exp0-7}}&
      \frame{\includegraphics[height=8.0ex]{imgs_sup/exp0-8}}&
      \frame{\includegraphics[height=8.0ex]{imgs_sup/exp0-9}}&
      \frame{\includegraphics[height=8.0ex]{imgs_sup/exp0-10}}&
      \frame{\includegraphics[height=8.0ex]{imgs_sup/exp0-n}} \\
      $\kappa_i=1.0$&
      0.8691&      
      0.8663&      
      &      
      &      
      &
      &
      &
      &
      ... \\
      \hline
      \multicolumn{1}{l}{}  &
      \multicolumn{3}{c}{\includegraphics[height=2.5ex]{imgs_sup/arrow_curve}} ${\scriptstyle 0.8691-0.8663>0.002}$&  
      &  
      &  
      &  
      &  
      & \multicolumn{1}{l}{} \\
      \multicolumn{10}{l}{\textbf{~~~$\cdots$}} \\
      \hline
      $I_\text{1}$&    
      $I_\text{2}$&      
      $I_\text{3}$&      
      $I_\text{4}$&      
      $I_\text{5}$&      
      Skipped&      
      &
      &
      &
      \\
      \frame{
        \begin{overpic}[height=8.0ex]{imgs_sup/exp0-0}
          \put (55,33) {\textcolor{ForestGreen}{\shadowtext{\huge\cmark}}}
        \end{overpic}
      }& 
      \frame{
        \begin{overpic}[height=8.0ex]{imgs_sup/exp0-3}
          \put (55,30) {\textcolor{ForestGreen}{\shadowtext{\huge\cmark}}}
        \end{overpic}
      }&
      \frame{
        \begin{overpic}[height=8.0ex]{imgs_sup/exp0-4}
          \put (55,20) {\textcolor{ForestGreen}{\shadowtext{\huge\cmark}}}
        \end{overpic}
      }&
      \frame{
        \begin{overpic}[height=8.0ex]{imgs_sup/exp0-5}
          \put (55,18) {\textcolor{ForestGreen}{\shadowtext{\huge\cmark}}}
        \end{overpic}
      }&
      \frame{
        \begin{overpic}[height=8.0ex]{imgs_sup/exp0-6}
          \put (55,22) {\textcolor{ForestGreen}{\shadowtext{\huge\cmark}}}
        \end{overpic}
      }&
      \frame{
        \begin{overpic}[height=8.0ex]{imgs_sup/exp0-7}
          \put (60,22) {\textcolor{BrickRed}{\shadowtext{\huge\xmark}}}
        \end{overpic}
      }&
      \frame{\includegraphics[height=8.0ex]{imgs_sup/exp0-8}}&
      \frame{\includegraphics[height=8.0ex]{imgs_sup/exp0-9}}&
      \frame{\includegraphics[height=8.0ex]{imgs_sup/exp0-10}}&
      \frame{\includegraphics[height=8.0ex]{imgs_sup/exp0-n}} \\
      $\kappa_i=1.0$&
      &      
      &      
      &      
      0.8490&      
      0.8488&
      &
      &
      &
      ... \\
      \hline
      \multicolumn{1}{l}{}  &
      &  
      &  
      &  
      \multicolumn{3}{c}{\includegraphics[height=2.5ex]{imgs_sup/arrow_curve}} ${\scriptstyle 0.8490-0.8488\leq0.002}$ &  
      &
      & \multicolumn{1}{l}{} \\
      \hline
      $I_\text{1}$&    
      $I_\text{2}$&      
      $I_\text{3}$&      
      $I_\text{4}$&      
      $I_\text{5}$&      
      Skipped&      
      $I_\text{6}$&
      &
      &
      \\
      \frame{
        \begin{overpic}[height=8.0ex]{imgs_sup/exp0-0}
          \put (55,33) {\textcolor{ForestGreen}{\shadowtext{\huge\cmark}}}
        \end{overpic}
      }& 
      \frame{
        \begin{overpic}[height=8.0ex]{imgs_sup/exp0-3}
          \put (55,30) {\textcolor{ForestGreen}{\shadowtext{\huge\cmark}}}
        \end{overpic}
      }&
      \frame{
        \begin{overpic}[height=8.0ex]{imgs_sup/exp0-4}
          \put (55,20) {\textcolor{ForestGreen}{\shadowtext{\huge\cmark}}}
        \end{overpic}
      }&
      \frame{
        \begin{overpic}[height=8.0ex]{imgs_sup/exp0-5}
          \put (55,18) {\textcolor{ForestGreen}{\shadowtext{\huge\cmark}}}
        \end{overpic}
      }&
      \frame{
        \begin{overpic}[height=8.0ex]{imgs_sup/exp0-6}
          \put (55,22) {\textcolor{ForestGreen}{\shadowtext{\huge\cmark}}}
        \end{overpic}
      }&
      \frame{
        \begin{overpic}[height=8.0ex]{imgs_sup/exp0-7}
          \put (60,22) {\textcolor{BrickRed}{\shadowtext{\huge\xmark}}}
        \end{overpic}
      }&
      \frame{
        \begin{overpic}[height=8.0ex]{imgs_sup/exp0-8}
          \put (55,25) {\textcolor{ForestGreen}{\shadowtext{\huge\cmark}}}
        \end{overpic}
      }&
      \frame{\includegraphics[height=8.0ex]{imgs_sup/exp0-9}}&
      \frame{\includegraphics[height=8.0ex]{imgs_sup/exp0-10}}&
      \frame{\includegraphics[height=8.0ex]{imgs_sup/exp0-n}} \\
      $\kappa_i=1.0$&
      &      
      &      
      &      
      0.8490&      
      &
      0.8456&
      &
      &
      ... \\
      \hline
      \multicolumn{1}{l}{}  &
      &  
      &  
      &  
      \multicolumn{5}{c}{\includegraphics[height=2.5ex]{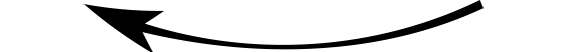}} ${\scriptstyle 0.8490-0.8456>0.002}$ \hspace{30pt} &  
      \multicolumn{1}{l}{} \\
    \multicolumn{10}{p{0.99\linewidth}}{} \\[0.5ex]
      \multicolumn{10}{p{0.99\linewidth}}{
        \textbf{(e)} Form an image subset $\mathcal{S}=\left\lbrace I_1,\ldots,I_6 \right\rbrace$ if 6 images can be greedily added (true for this example), otherwise discard the entire set and restart at \textbf{(a)}.
      }
    \end{tabular}
    \end{minipage}
    \\[4pt]
    \begin{minipage}{0.98\linewidth}
    \centering
    \setlength{\tabcolsep}{2pt}
    \begin{tabular}{|cccccc|}
    \hline
    $I_\text{1}$&    
    $I_\text{2}$&    
    $I_\text{3}$&    
    $I_\text{4}$&    
    $I_\text{5}$&    
    $I_\text{6}$
    \\
    \frame{\includegraphics[height=8.0ex]{imgs_sup/exp0-0}}& 
    \frame{\includegraphics[height=8.0ex]{imgs_sup/exp0-3}}&
    \frame{\includegraphics[height=8.0ex]{imgs_sup/exp0-4}}&
    \frame{\includegraphics[height=8.0ex]{imgs_sup/exp0-5}}&
    \frame{\includegraphics[height=8.0ex]{imgs_sup/exp0-6}}&
    \frame{\includegraphics[height=8.0ex]{imgs_sup/exp0-8}}
    \\
    \hline
  \end{tabular}
\end{minipage}
  \caption{The procedure of forming an image subset as described in \secref{sec:dset_imgset}. We specifically show cases where images are removed or skipped. Note that after forming the subsets, we further filter them to avoid heavy overlaps.}
  \label{fig:sup-0}
\end{figure*}

\figref{fig:sup-0} shows the procedure for constructing an image subset of six elements, noted as $\mathcal{S}=\left\lbrace I_1,\ldots,I_6 \right\rbrace$ in \secref{sec:dset_imgset}. We specifically demonstrate cases where images are removed. The process was designed to ensure that images in a given subset are visually similar to one another while exhibiting some appreciable differences.

\paragraph{Image pairs.}
\shadowoffset{2pt}
\begin{figure*}[!ht]
  \centering\footnotesize
  \noindent
  \begin{minipage}{.25\textwidth}
    \centering
    \includegraphics[height=20.0ex]{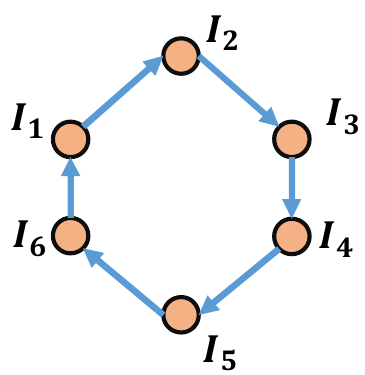}
  \end{minipage}
  \begin{minipage}{.75\textwidth}
    \centering\setlength{\tabcolsep}{2pt}
    \begin{tabular}{cccccc}
      $I_\text{1}$& $I_\text{2}$ & $I_\text{3}$ & $I_\text{4}$ & $I_\text{5}$ & $I_\text{6}$         \\
      \frame{\includegraphics[height=12.0ex]{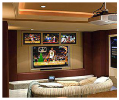}}&  
      \frame{\includegraphics[height=12.0ex]{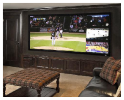}}&  
      \frame{\includegraphics[height=12.0ex]{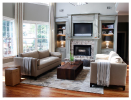}}&  
      \frame{\includegraphics[height=12.0ex]{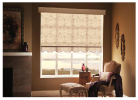}}&  
      \frame{\includegraphics[height=12.0ex]{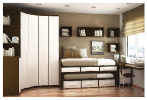}}&          
      \frame{\includegraphics[height=12.0ex]{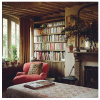}}\\ 
      \\[0.5ex]
      \multicolumn{6}{p{0.85\linewidth}}{
        $I_\text{1}\rightarrow I_\text{2}$: Turn on the flat screen tv in the living room.
      }    \\
      \multicolumn{6}{p{0.85\linewidth}}{
        $I_\text{2}\rightarrow I_\text{3}$: Pull up the blinds to let in sunlight.
      }    \\
      \multicolumn{6}{p{0.85\linewidth}}{
        $I_\text{3}\rightarrow I_\text{4}$: Put a window by the fireplace.
      }    \\
      \multicolumn{6}{p{0.85\linewidth}}{
        $I_\text{4}\rightarrow I_\text{5}$: Have a window on the right-hand wall.
      }    \\
      \multicolumn{6}{p{0.85\linewidth}}{
        $I_\text{5}\rightarrow I_\text{6}$: Have a bookshelf to the right of the window.
      }    \\
      \multicolumn{6}{p{0.85\linewidth}}{
        $I_\text{6}\rightarrow I_\text{1}$: Have multiple television screens.
      }    \\
      \multicolumn{6}{l}{}   
      \end{tabular}
  \end{minipage}
  \caption{Left: The six pairs we draw from a subset (in total we draw nine) that form a closed-loop dialogue. Each arrow represents a reference-to-target image pair with modification sentences.
  Right: An example of consecutive modification sentences that forms a dialogue.}
  \label{fig:dialogue_0}
\end{figure*}

\begin{figure*}[!ht]
  \centering\footnotesize
  \noindent
  \begin{minipage}{.25\textwidth}
    \centering
    \includegraphics[height=20.0ex]{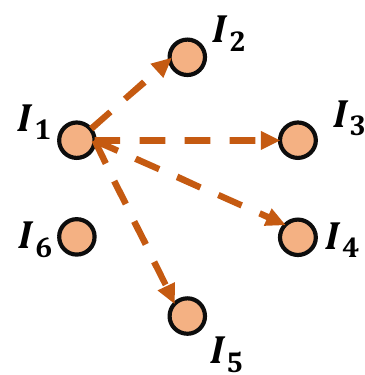}
  \end{minipage}
  \begin{minipage}{.75\textwidth}
    \centering\setlength{\tabcolsep}{2pt}
    \begin{tabular}{cccccc}
      $I_\text{1}$& $I_\text{2}$ & $I_\text{3}$ & $I_\text{4}$ & $I_\text{5}$ & $I_\text{6}$         \\
      \frame{\includegraphics[height=12.0ex]{imgs_sup/exp1-0-train-4025-0-img0}}&  
      \frame{\includegraphics[height=12.0ex]{imgs_sup/exp1-5-train-11850-1-img1}}&  
      \frame{\includegraphics[height=12.0ex]{imgs_sup/exp1-4-train-2379-1-img1}}&  
      \frame{\includegraphics[height=12.0ex]{imgs_sup/exp1-3-train-7125-1-img1}}&  
      \frame{\includegraphics[height=12.0ex]{imgs_sup/exp1-2-train-7537-3-img0}}&          
      \frame{\includegraphics[height=12.0ex]{imgs_sup/exp1-1-train-10458-2-img1}}\\ 
      \\[0.5ex]
      \multicolumn{6}{p{0.85\linewidth}}{
        $I_\text{1}\rightarrow I_\text{2}$: Turn on the flat screen tv in the living room.
      }    \\
      \multicolumn{6}{p{0.85\linewidth}}{
        $I_\text{1}\rightarrow I_\text{3}$: A hall with two bright sofas and a brown table between them.
      }    \\
      \multicolumn{6}{p{0.85\linewidth}}{
        $I_\text{1}\rightarrow I_\text{4}$: Room with a large window, a bright armchair and a fireplace.
      }    \\
      \multicolumn{6}{p{0.85\linewidth}}{
        $I_\text{1}\rightarrow I_\text{5}$: A room with a large window and windowsill, a high sofa and shelves above it, instead of a hall with one large sofa and four screens in front of it.
      }    \\
      \multicolumn{6}{l}{}   
      \end{tabular}
  \end{minipage}
  \caption{Left: The four pairs we draw from a subset to have multiple outcomes from the same reference image. Each arrow represents a reference-to-target image pair with modification sentences.
  Right: An example of the four pairs with the same reference image.}
  \label{fig:dialogue_1}
\end{figure*}

\begin{figure*}[!ht]
  \centering\footnotesize
  \noindent
  \begin{minipage}{.33\textwidth}
    \centering
    \includegraphics[height=20.0ex]{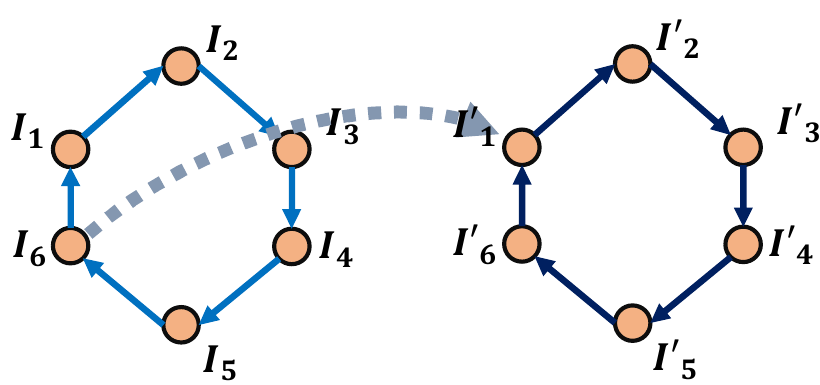}
  \end{minipage}  
  \begin{minipage}{.65\textwidth}
    \centering\setlength{\tabcolsep}{2pt}
    \begin{tabular}{cccccc}
      $I_\text{1}$& $I_\text{2}$ & $I_\text{3}$ & $I_\text{4}$ & $I_\text{5}$ & $I_\text{6}$         \\
      \frame{\includegraphics[height=10.0ex]{imgs_sup/exp1-0-train-4025-0-img0}}&  
      \frame{\includegraphics[height=10.0ex]{imgs_sup/exp1-5-train-11850-1-img1}}&  
      \frame{\includegraphics[height=10.0ex]{imgs_sup/exp1-4-train-2379-1-img1}}&  
      \frame{\includegraphics[height=10.0ex]{imgs_sup/exp1-3-train-7125-1-img1}}&  
      \frame{\includegraphics[height=10.0ex]{imgs_sup/exp1-2-train-7537-3-img0}}&          
      \frame{\includegraphics[height=10.0ex]{imgs_sup/exp1-1-train-10458-2-img1}}\\ 
      \\
      $I'_\text{1}$& $I'_\text{2}$ & $I'_\text{3}$ & $I'_\text{4}$ & $I'_\text{5}$ & $I'_\text{6}$         \\
      \frame{\includegraphics[height=10.0ex]{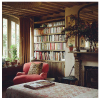}}&  
      \frame{\includegraphics[height=10.0ex]{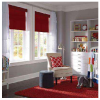}}&  
      \frame{\includegraphics[height=10.0ex]{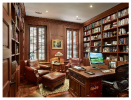}}&  
      \frame{\includegraphics[height=10.0ex]{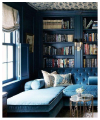}}&  
      \frame{\includegraphics[height=10.0ex]{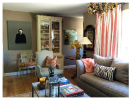}}&          
      \frame{\includegraphics[height=10.0ex]{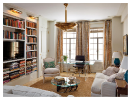}}\\ 
      \multicolumn{6}{l}{}   
      \end{tabular}
  \end{minipage}
  \caption{An example of connecting pairs from two subsets to form longer dialogue paths. Note that in this example, $I_\text{6} \equiv I'_\text{1}$.}
  \label{fig:dialogue_2}
\end{figure*}

As explained in \secref{sec:dset_imgset}, we draw nine pairs from each subset. \figref{fig:dialogue_0} demonstrates how we form consecutive modifications among pairs, which could facilitate the training and evaluation of dialogue systems in the future.
\figref{fig:dialogue_1} shows that one reference image leads to multiple targets in each subset. This should allow the study of the impact of language modality in the future.

We point out that the length of dialogue paths can vary for two reasons.
First, we allow a slight overlap between the images of two subsets. Therefore, it is possible to form dialogue paths across subsets with variable lengths, as shown in \figref{fig:dialogue_2}.
Second, AMT workers can mark pairs of poor quality and choose not to annotate them (see below). Such pairs will be removed from the dataset, thus rendering the dialogue incomplete. In total, 71.1\% of the subsets have closed-loop dialogue paths (see \tabref{tab:dst_stats_ext} for detailed statistics).

\paragraph{Annotation collection on AMT.}
\tabref{tab:guideline_0} demonstrates our guideline to AMT workers, specifying types of annotations to avoid. 

\begin{figure*}[!ht]
  \centering\footnotesize
  \noindent
  \begin{minipage}{.95\textwidth}
    \centering
    \fbox{
    \includegraphics[width=95.0ex]{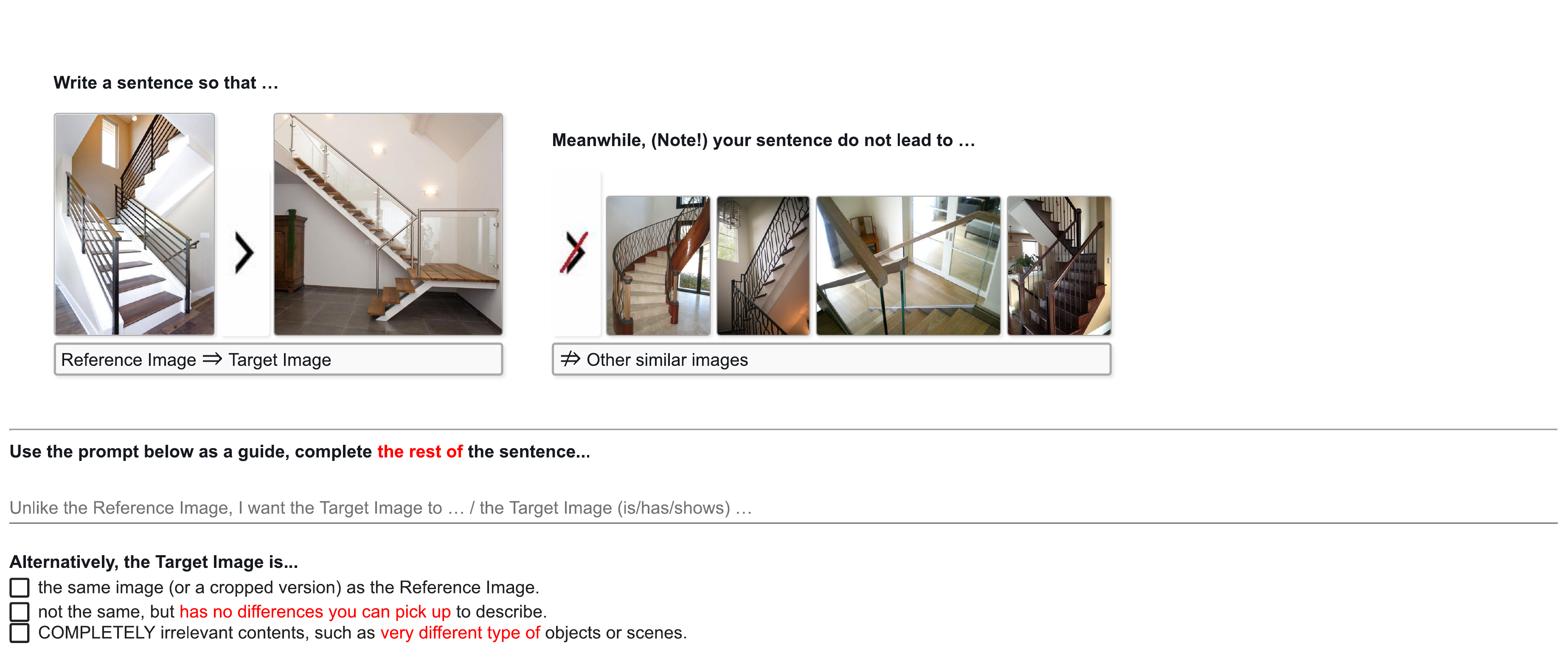}
    }
  \end{minipage}
  \\[3pt]
  \begin{minipage}{.95\textwidth}
    \centering
    \fbox{
    \includegraphics[width=95.0ex]{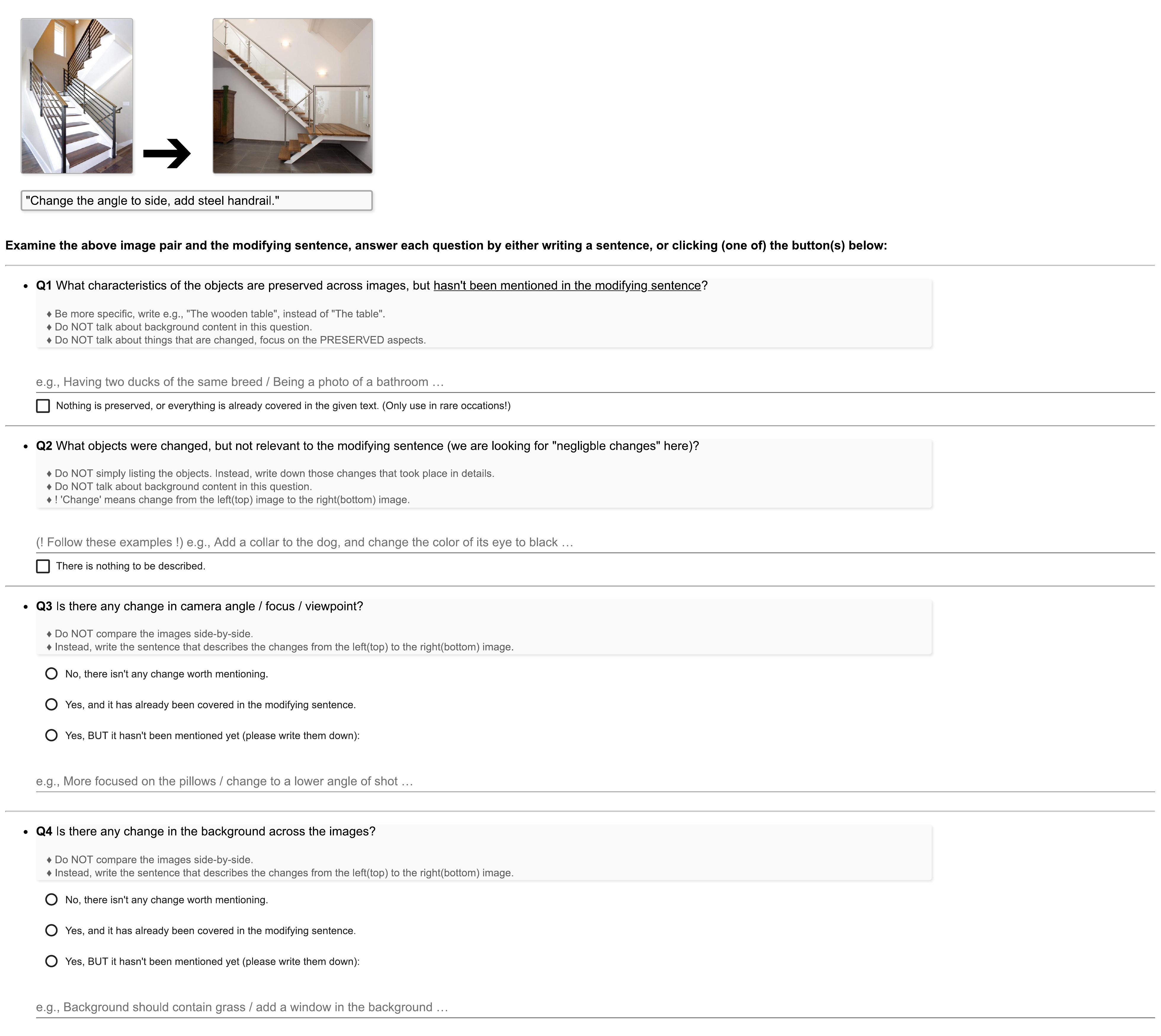}
    }
  \end{minipage}
  \\[3pt]
  \begin{minipage}{.95\textwidth}
    \centering
    \fbox{
    \includegraphics[width=95.0ex]{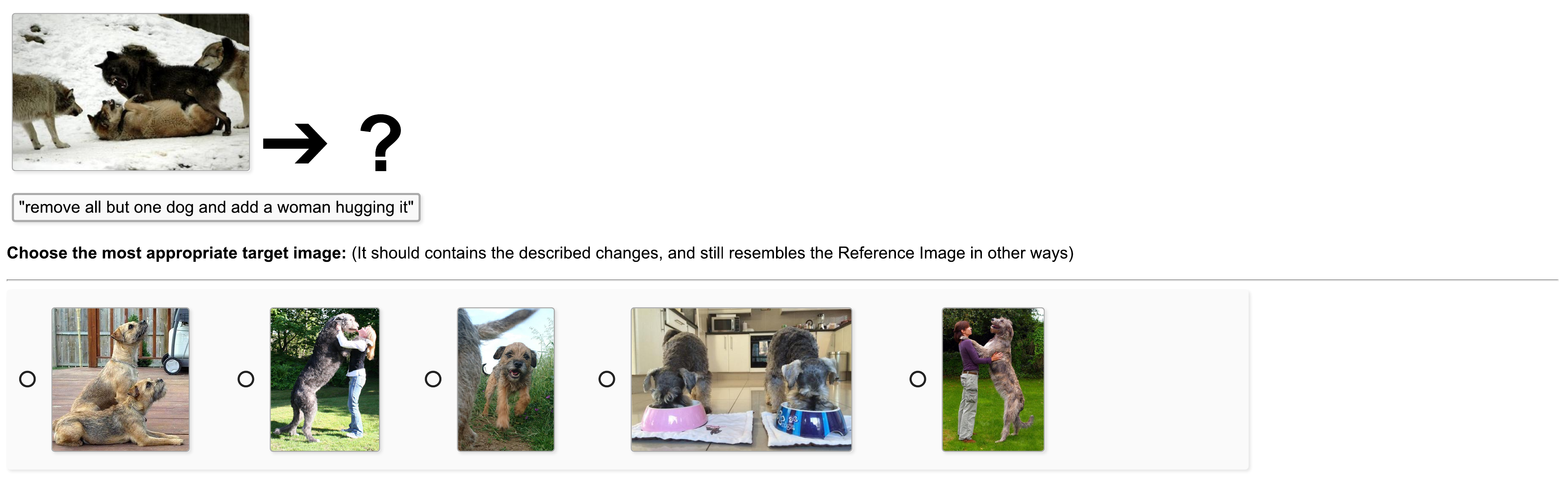}
    }
  \end{minipage}
  
  \caption{Snapshots of our collection interface (recommend viewing digitally by zooming in). (top) (Main) annotation, where we specifically require unique sentences within each subset. (middle) Auxiliary annotation, where we ask Amazon Mechanical Turk (AMT) workers of four detailed questions per pair. (bottom) Human performance (Recall$_{\text{Subset}}@1$) evaluation on test-split, where we ask AMT workers to choose the most probable target image within the subset.}
  \label{fig:amt_interface}
\end{figure*}

\figref{fig:amt_interface} shows our collection interface. (top) For main annotations, we require AMT workers to write sentences that only lead to the true target image, thus removing false-negatives in each subset. We also allow them to mark image pairs of poor quality for removal. (middle) For auxiliary annotations, we ask four detailed questions to clarify ambiguities within the given pair. (bottom) We evaluate human retrieval performance in Recall$_{\text{Subset}}$ using the test-split.

\paragraph{Quality control.}
We conduct a pre-selection process to manually whitelist workers with good annotation quality. Our pre-selection procedure plays a critical role in quality assurance, where we filter out over 60\% of the submitted workers. Workers who have passed the selection process produce annotations with over 90\% acceptance rate.

For annotations submitted by workers in the whitelist, we manually review $\sim$30\% of the annotations from each worker. The remaining are examined with an automated script to check for potential abuse of the use of checkboxes, irresponsible contents (\eg, very short sentences), and annotations that violate our guidelines.


\paragraph{Human performance.}
\tabref{tab:baseline_0} (row 7) lists the human retrieval performance of Recall$_{\text{Subset}}@1$ (see \secref{sec:extended_metric} for details of the Recall$_{\text{Subset}}$ metric) on test-split. Here, we present the collection procedures of this score.

\figref{fig:amt_interface} (bottom) shows the collection interface. Specifically, we ask AMT workers to choose the most probable target image for a given text-image query.
We employ three different AMT workers for each pair in the test-split, our final score is calculated by averaging over all submitted results.

\section{Additional Analysis on \dstname}
\begin{sidewaysfigure*}[!ht]
  \centering
    \includegraphics[trim={0 1pt 0 0},clip, width=0.95\linewidth]{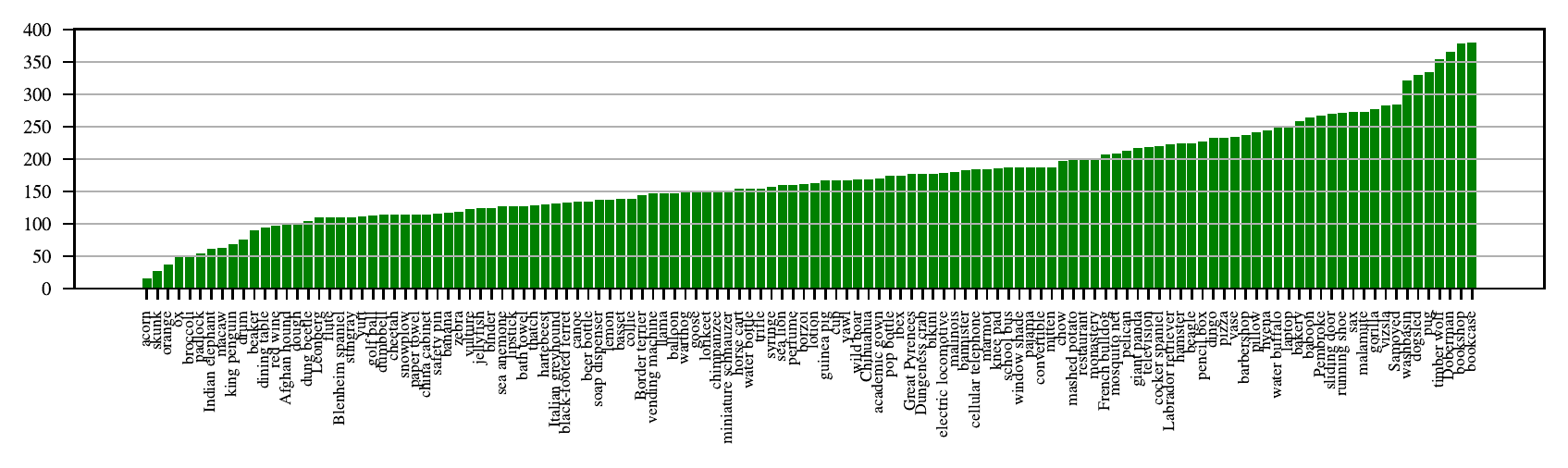}
  \caption{Number of examples per synset (sorted in ascending).}
  \label{fig:synset}
\end{sidewaysfigure*}

\paragraph{Image synsets.}
We analyze the image contents using the synset information in NLVR$^2$~\cite{Suhr_2019_nlvr2}. CIRR includes 124 out of the 1000 synsets in NLVR$^2$. Each synset is associated with 136.6$\pm$73.1 ($\mu\pm\sigma$) images.
The five most common synsets are \texttt{bookcase}, \texttt{bookshop}, \texttt{dobreman}, \texttt{timber wolf} and \texttt{pug}. The five least common synsets are \texttt{acorn}, \texttt{skunk}, \texttt{orange}, \texttt{ox}, \texttt{broccoli} and \texttt{padlock}. Distributions of samples are shown in \figref{fig:synset}.

Note that we do not distinguish synsets of similar concepts (\eg, \texttt{dobreman} and \texttt{French bulldog}) when forming image pairs, instead, we choose by visual similarity. Additionally, we point out that for composed image retrieval, synset may not fully characterize an image, as the annotations focus on fine-grained visual comparisons.

\paragraph{Comparison to existing datasets.}

\newcolumntype{L}{>{\centering\arraybackslash}m{3cm}}
\begin{table*}
  \begin{minipage}{.98\textwidth}
  \centering \scalebox{0.7}{
  \renewcommand{\arraystretch}{1.05} 
    \centering
    \begin{tabular}{llp{0.05\linewidth}p{0.10\linewidth}p{0.10\linewidth}p{0.08\linewidth}p{0.08\linewidth}p{0.08\linewidth}p{0.08\linewidth}p{0.18\linewidth}lp{0.09\linewidth}} 
      \toprule
        & \multirow{2}{*}{Datasets} & \multicolumn{2}{c}{\phantom{ttttttttttttttttt}Statistics} & \multicolumn{3}{c}{\phantom{ttttttttttttttttttttttttttttttttttttt}Images}  & \multicolumn{3}{c}{\phantom{tttttttttttt}Annotations}                                                                                                                                                    & \multirow{2}{0.09\linewidth}{Non-repurposed} & \multirow{2}{0.09\linewidth}{Additional annotation}          \\ 
      \cmidrule(lr){3-5}
      \cmidrule(lr){6-8}
      \cmidrule(lr){9-10}
        &                           & Nb. pairs & Nb. images          & Domain               & Natural images & Pairing strategy      & Natural language & Dialogue paths & Examples                                                                                                                                        &                                 &                                                 \\ 
      \midrule
      1 & CSS~\cite{Vo_2019_tirg}   & --        & 38,069$^\ast$              & --                   &                & --                    &                  &                & \textit{Make top-center blue object yellow.}                                                                                                             & \cmark                           &                                                 \\
      \midrule
      2 & MIT-States~\cite{Vo_2019_tirg,Isola2015DiscoveringSA_mitstates}                & --        & 63,440$^{\ast}$              & Entity states        & \cmark          & by entity class       &                  &                & \textit{(Change state) to melted.}                                                                                                                       &                                 &                                                 \\
      \midrule
      3 & Fashion200k~\cite{Vo_2019_tirg,han2017automatic_fashion200k}               & --        & 338,372$^{\ast~\dagger}$             & Fashion              & \cmark          & by similar attributes &                  &                & \textit{(Replace with) beige.}                                                                                                                           &                                 &                                                 \\
      \midrule
      4 & Fashion-IQ~\cite{fashioniq}                & 30,122$^{\S}$    & 46,609$^\ddagger$              & Fashion              & \cmark          & by product category   & \cmark            &                & \textit{Is short sleeved and has a collar.}                                                                                                              & \cmark                           & Product attribute (partial)                               \\
      \midrule
      5 & Birds-to-Words~\cite{dodds2020modality_maaf,forbes2019neural_birds}            & 3,347     & --                  & Birds                & \cmark          & by visual similarity  & \cmark            &                & \textit{Animal1 is white with dark brown and
      white wings and a golden head . Animal2
      is brown-gold with dark solid-colored
      brown wings and a dark head.} &                                 &                                                 \\
      \midrule
      6 & Spot-the-Diff~\cite{dodds2020modality_maaf,jhamtani2018learning_spotthediff}             & 23,089    & --                  & Surveillance footage & \cmark          & by video frame        & \cmark            &                & \textit{A white truck has appeared in the after image. A person is now walking on the footpath.}                                                         &                                 &                                                 \\ 
      \midrule
      7 & \textbf{CIRR}                      & \textbf{36,554}    & \textbf{21,552}              & \textbf{Open}                 & \cmark          & \textbf{by visual similarity}  & \cmark            & \cmark          & \textit{\textbf{Room with a large window, a bright armchair and a fireplace.}}                                                                                    & \cmark                           & \textbf{Auxiliary annotation clarifying ambiguities}  \\
      \bottomrule
    \end{tabular}}
\end{minipage}\\[1pt]
\begin{minipage}{.98\textwidth}
  \fontsize{7pt}{8pt}\selectfont
  $\ast$ Nb. pairs not pre-defined, pairs are generated on-the-fly.\\[1.75pt]
  $\dagger$ Approx. 100,000 images have low detection score, thus could be removed~\cite{han2017automatic_fashion200k}. Here, we show the available nb. images in total.\\[1.75pt]
  $\S$ Each pair has two sentences.\\[1.75pt]
  $\ddagger$ Combining all three subtypes. Note that pairs and images overlap between subtypes.\\[1.75pt]
\end{minipage}
  \caption{Comparison between CIRR (bolded) and existing datasets for composed image retrieval. CIRR is comparable in size (nb. pairs) while containing richer annotations of open-domain images.}
  \label{tab:comparison_dst}
  \end{table*}

\tabref{tab:comparison_dst} compares CIRR with existing datasets used for composed image retrieval. We demonstrate that CIRR is comparable in size with existing datasets. Additionally, it provides rich auxiliary annotations for open-domain images.

\paragraph{False-negative analysis.}
\shadowoffset{2pt}
\begin{figure*}[!ht]
  \centering\footnotesize
  \begin{minipage}{0.98\linewidth}
    \centering
    \setlength{\tabcolsep}{3pt}
    \begin{tabular}{lcccccc}
      \textbf{(a)}&
      \multicolumn{6}{l}{Is shiny and silver with shorter sleeves $+$ fit and flare.}  
      \\
      \frame{\includegraphics[height=15.0ex]{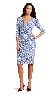}}& 
      \frame{\includegraphics[height=15.0ex]{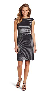}}& 
      \frame{\includegraphics[height=15.0ex]{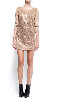}}&
      \frame{\includegraphics[height=15.0ex]{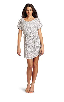}}&
      \frame{\includegraphics[height=15.0ex]{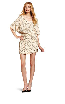}}&
      \frame{\includegraphics[height=15.0ex]{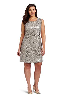}}&
      \frame{\includegraphics[height=15.0ex]{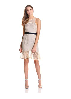}}\\
      \multicolumn{7}{l}{} \\[0.05ex]

      \textbf{(b)}&
      \multicolumn{6}{l}{Is less formal with different colored stripes $+$ Does not have a collar.}  
      \\[1ex]
      \frame{\includegraphics[height=15.0ex]{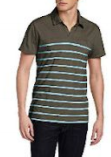}}& 
      \frame{\includegraphics[height=15.0ex]{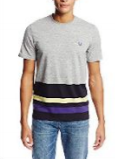}}& 
      \frame{\includegraphics[height=15.0ex]{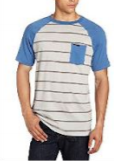}}&
      \frame{\includegraphics[height=15.0ex]{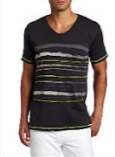}}&
      \frame{\includegraphics[height=15.0ex]{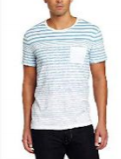}}&
      \frame{\includegraphics[height=15.0ex]{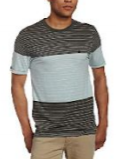}}&
      \frame{\includegraphics[height=15.0ex]{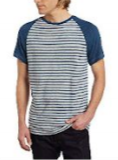}}\\
      \multicolumn{7}{l}{} \\[0.05ex]

      \textbf{(c)}&
      \multicolumn{6}{l}{Is a solid black color, also shorter and tighter fitting $+$ Is black and more skimpy.}  
      \\[1ex]
      \frame{\includegraphics[height=15.0ex]{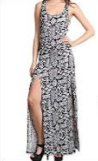}}& 
      \frame{\includegraphics[height=15.0ex]{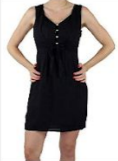}}& 
      \frame{\includegraphics[height=15.0ex]{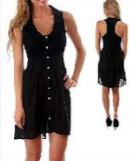}}&
      \frame{\includegraphics[height=15.0ex]{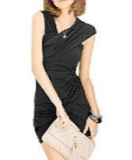}}&
      \frame{\includegraphics[height=15.0ex]{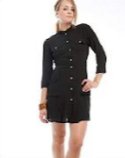}}&
      \frame{\includegraphics[height=15.0ex]{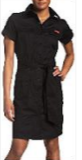}}&
      \frame{\includegraphics[height=15.0ex]{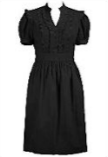}}\\
      \multicolumn{7}{l}{} \\[0.05ex]

      \textbf{(d)}&
      \multicolumn{6}{l}{Has more grey and longer sleeves $+$ Is lighter.}  
      \\[1ex]
      \frame{\includegraphics[height=15.0ex]{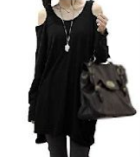}}& 
      \frame{\includegraphics[height=15.0ex]{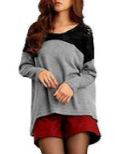}}& 
      \frame{\includegraphics[height=15.0ex]{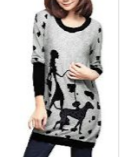}}&
      \frame{\includegraphics[height=15.0ex]{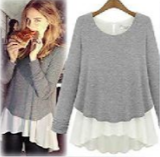}}&
      \frame{\includegraphics[height=15.0ex]{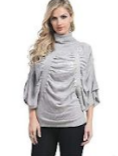}}&
      \frame{\includegraphics[height=15.0ex]{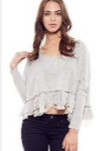}}&
      \frame{\includegraphics[height=15.0ex]{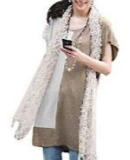}}\\
    \end{tabular}
    \end{minipage}
  \caption{Examples of false-negatives in Fashion-IQ~\cite{fashioniq}. First column shows the reference image. Each sample contains two modification sentences. For each query set (reference image + modification sentences), only one candidate image is labeled as the target. Thus, rendering the remaining valid predictions as false-negatives.}
  \label{fig:false-negatives}
\end{figure*}

\figref{fig:false-negatives} demonstrates the presence of false-negatives in Fashion-IQ~\cite{fashioniq}, as explained in \secref{dataset}. 
For comparison, our data collection procedures ensure that no false-negatives are present within each image subset, as discussed in \secref{sec:sup_subset}. Examples of CIRR are shown in \figref{fig:dialogue_0}, \figref{fig:dialogue_1}, and \figref{fig:cirr_examples}.

\section{Additional Examples of \dstname} \label{sec:sup_examples}
\shadowoffset{2pt}
\setlength{\fboxsep}{0pt}
\begin{figure*}[!ht]
  \centering\footnotesize
  \begin{minipage}{0.98\linewidth}
    \centering
    \setlength{\tabcolsep}{1.5pt}
    \begin{tabular}{lccccc}
      \textbf{(a)}&
      \\
      \frame{\includegraphics[height=12.0ex]{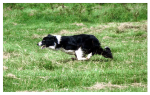}}& 
      \frame{\includegraphics[height=12.0ex]{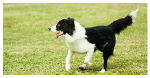}}& 
      \frame{\includegraphics[height=12.0ex]{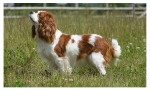}}&
      \frame{\includegraphics[height=12.0ex]{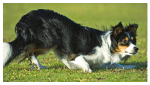}}&
      \frame{\includegraphics[height=12.0ex]{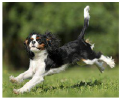}}&
      \textcolor{ForestGreen}{\fboxrule=2pt\fbox{\includegraphics[height=12.0ex]{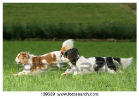}}}\\
      &\multicolumn{5}{l}{
        \textit{Main} -- Goes from a black and white dog running to two dogs running.
      }  \\
      &\multicolumn{5}{l}{
        \textit{Q1} -- [N/A] Nothing worth mentioning
      }\\
      &\multicolumn{5}{l}{
        \textit{Q2} -- \textbf{Change to a brown-and-white dog and a black-and-white dog.}
      }\\
      &\multicolumn{5}{l}{
        \textit{Q3} -- [N/A] Nothing worth mentioning
      }\\
      &\multicolumn{5}{l}{
        \textit{Q4} -- Make the grass a darer green.
      }\\[0.05ex]
      \multicolumn{6}{l}{}\\
      \textbf{(b)}&
      \\
      \frame{\includegraphics[height=12.0ex]{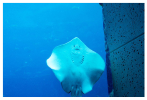}}& 
      \frame{\includegraphics[height=12.0ex]{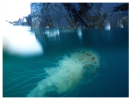}}& 
      \frame{\includegraphics[height=12.0ex]{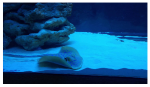}}&
      \frame{\includegraphics[height=12.0ex]{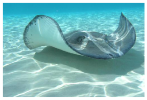}}&
      \textcolor{ForestGreen}{\fboxrule=2pt\fbox{\includegraphics[height=12.0ex]{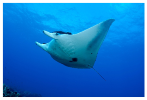}}}&
      \frame{\includegraphics[height=12.0ex]{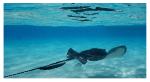}}\\
      &\multicolumn{5}{l}{
        \textit{Main} -- Remove the concret to the right.
      }  \\
      &\multicolumn{5}{l}{
        \textit{Q1} -- \textbf{Has marine animal in similar blue backdrop.}
      }\\
      &\multicolumn{5}{l}{
        \textit{Q2} -- Remove the blue thing on right.
      }\\
      &\multicolumn{5}{l}{
        \textit{Q3} -- [N/A] Nothing worth mentioning
      }\\
      &\multicolumn{5}{l}{
        \textit{Q4} -- [N/A] Nothing worth mentioning
      }\\[0.05ex]
      \multicolumn{6}{l}{}\\
      \textbf{(c)}&
      \\
      \frame{\includegraphics[height=12.0ex]{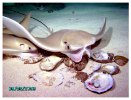}}& 
      \frame{\includegraphics[height=12.0ex]{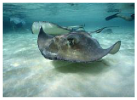}}& 
      \frame{\includegraphics[height=12.0ex]{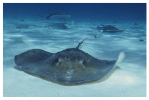}}&
      \textcolor{ForestGreen}{\fboxrule=2pt\fbox{\includegraphics[height=12.0ex]{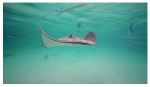}}}&
      \frame{\includegraphics[height=12.0ex]{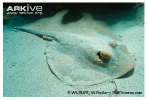}}&
      \frame{\includegraphics[height=12.0ex]{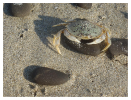}}\\
      &\multicolumn{5}{l}{
        \textit{Main} -- Remove the seashells and make the water green.
      }  \\
      &\multicolumn{5}{l}{
        \textit{Q1} -- Shows manta rays.
      }\\
      &\multicolumn{5}{l}{
        \textit{Q2} -- Make the rays older, spread the rays further apart.
      }\\
      &\multicolumn{5}{l}{
        \textit{Q3} -- \textbf{View straight on.}
      }\\
      &\multicolumn{5}{l}{
        \textit{Q4} -- [N/A] Covered in main annotation
      }\\[0.05ex]
      \multicolumn{6}{l}{}\\
      \textbf{(d)}&
      \\
      \frame{\includegraphics[height=12.0ex]{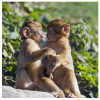}}& 
      \frame{\includegraphics[height=12.0ex]{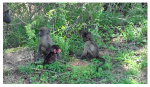}}& 
      \textcolor{ForestGreen}{\fboxrule=2pt\fbox{\includegraphics[height=12.0ex]{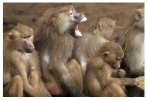}}}&
      \frame{\includegraphics[height=12.0ex]{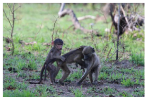}}&
      \frame{\includegraphics[height=12.0ex]{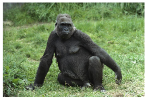}}&
      \frame{\includegraphics[height=12.0ex]{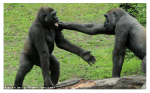}}\\
      &\multicolumn{5}{l}{
        \textit{Main} -- More monkeys
      }  \\
      &\multicolumn{5}{l}{
        \textit{Q1} -- \textbf{A group of monkeys side by side in same color.}
      }\\
      &\multicolumn{5}{l}{
        \textit{Q2} -- [N/A] Nothing worth mentioning
      }\\
      &\multicolumn{5}{l}{
        \textit{Q3} -- More focused on the animals.
      }\\
      &\multicolumn{5}{l}{
        \textit{Q4} -- [N/A] Nothing worth mentioning
      }\\[0.05ex]
    \end{tabular}
    \end{minipage}\\[2pt]
  \caption{Negative results of TIRG and \modelname on \dstname. Here, we show the Recall$_{\text{Subset}}$ rankings where we consider candidates from corresponding image subsets (see \secref{sec:extended_metric}). First column shows the reference images. True targets are in green boxes. Each pair contains a main annotation (Main) and four auxiliary annotations (Q1--4) as explained in \secref{sec:dset_annotations} and \secref{sec:aux_anno}.  We demonstrate errors of the models where: (a) fails to associate text with both reference and target image; and (b--d) fails to identify and preserve implicit global visual similarity. We show that \dstname focuses on the challenging task of distinguishing harder negatives that require fine-grained visual reasoning. Let us note that the errors can be explicitly interpreted with our auxiliary annotations (bolded), which previous datasets cannot. This suggests that future work can leverage the auxiliary annotations for analysis of methods, and possibly training of models that account for implicit human ambiguities.}
  \label{fig:cirr_examples}
\end{figure*}

We provide additional examples from the dataset in \figref{fig:cirr_examples}, where we demonstrate negative retrieval results from both TIRG and \modelname. 
We point out that \dstname focuses more on the challenging task of distinguishing among visually similar images.
Let us note that the auxiliary annotations provide explicit interpretations of errors, particularly regarding the implicit human-agreements in visual and language modalities. This suggests that the annotations can be used for fine-grained analysis or as training signals in future research on composed image retrieval.

\begin{table*}
  \centering \scalebox{0.71}{
    \centering
    \renewcommand{\arraystretch}{1.05} 
    \begin{tabular}{lll}
    \toprule
      & To avoid                                                                       & Examples  \\
    \midrule
    1 & Mentioning text/numbers                                                        & \textit{Text on the pillow says ``LOVE''.}          \\
    2 & Discussing photo editing properties                                            & \textit{Crop the staircases out of the photo.}          \\
    3 & Subjective opinions                                                            & \textit{The dogs look very cute.}          \\
    \midrule
    4 & Simply describing the target image, not comparing the pair                     &  \textit{Having a large table in the center of the room.}         \\
    5 & Simple side-by-side comparison                                                 &  \textit{The left image shows a laptop on the wooden table, the right image has a flatscreen.}         \\
    \midrule
    6 & Writing sentences that are not unique for the given image pair in the subset   &  --        \\
    \bottomrule
  \end{tabular}}
  \caption{Types of annotations we discourage workers from writing. Rows 4 and 5 might be admissible if the annotation contains implicit comparisons. (see \figref{fig:amt_interface} (top)).}
  \label{tab:guideline_0}
\end{table*}

\section{Dataset File Description}

  \begin{table*}
  \begin{minipage}{.98\textwidth}
  \centering \scalebox{0.7}{
  \renewcommand{\arraystretch}{1.05} 
    \centering
    \begin{tabular}{llp{0.25\linewidth}p{0.45\linewidth}p{0.38\linewidth}} 
      \toprule
         & Identifiers (keys)              & Explanations                                                           & Content Details (values)                                                                                     & Examples                                                                                              \\ 
      \midrule
      1  & \texttt{pairid}          & Unique pair id$^\ast$                                                  &                                                                                                      & \texttt{12554}                                                                                                 \\ 
      \midrule
      2  & \texttt{reference}       & Reference image                                                        & \multirow{3}{0.98\linewidth}{Follow NLVR$^2$~\cite{Suhr_2019_nlvr2} image naming conventions.}       & \texttt{"dev-147-2-img0"}                                                                                      \\
      3  & \texttt{target\_hard}    & Target image$^{\S}$                                                           &                                                                                                      & \texttt{"dev-846-2-img0"}                                                                                      \\
      4  & \texttt{target\_soft}    & Target image with additional labeling (if exists)$^{\S}$$^\dagger$            &                                                                                                      & \texttt{\{dev-846-2-img0": 1.0, "dev-743-3-img0": -1.0\}}                                                      \\ 
      \midrule
      5  & \texttt{caption}         & (Main) annotation                                                       &                                                                                                      & \texttt{"Catch the crab in the circular ring and place them on the metal table."}                              \\ 
      \midrule
      6  & \texttt{caption\_extend} & \multicolumn{1}{l}{Auxiliary annotation$^\ddagger$}                            & \multicolumn{1}{l}{}                                                                                 &                                                                                                       \\ 
      \cmidrule{2-5}
      7  & \texttt{0}               & Q1                                                                     & \multirow{2}{0.98\linewidth}{Begin with \texttt{[c]} if N/A.}                                        & \texttt{"[c] None existed"}                                                                                    \\
      8  & \texttt{1}               & Q2                                                                     &                                                                                                      & \texttt{"We don't see the gloved hands of the fisherman"}                                                      \\
      \cmidrule{2-5}
      9  & \texttt{2}               & Q3                                                                     & \multirow{2}{0.98\linewidth}{Begin with \texttt{[cr0]} if Nothing worth mentioning, begin with \texttt{[cr1]} if Covered in brief annotation.} & \texttt{"Focus on the net full of crabs"}                                                         \\
      10 & \texttt{3}               & Q4                                                                     &                                                                                                      & \texttt{"[cr0] Nothing worth mentioning"}                                                                      \\ 
      \midrule
      11 & \texttt{img\_set}        & \multicolumn{1}{l}{Subset information}                                 & \multicolumn{1}{l}{}                                                                                 &                                                                                                       \\ 
      \cmidrule{2-5}
      12 & \texttt{id}              & Unique subset id                                                       &                                                                                                      & \texttt{106}                                                                                                   \\
      \cmidrule{2-5}
      13 & \texttt{members}         & Images within subset                                                   & Follow NLVR$^2$~\cite{Suhr_2019_nlvr2} image naming conventions.                                     & \texttt{["dev-147-2-img0", "dev-224-1-img1", "dev-410-2-img0", "dev-743-3-img0", "dev-846-2-img0", "dev-998-1-img0"]}  \\ 
      \cmidrule{2-5}
      14 & \texttt{reference\_rank} & \multirow{2}{0.70\linewidth}{Sequence identifier as in \figref{fig:dialogue_0}$^{\S}$ } & \multirow{2}{0.98\linewidth}{Range from 0 to 5, correspond to $I_\text{1}$-$I_\text{6}$.}  & \texttt{0}                                                                                                     \\
      15 & \texttt{target\_rank}    &                                                                        &                                                                                                      & \texttt{1}                                                                                                     \\
      \bottomrule
    \end{tabular}}
\end{minipage}\\[1pt]
\begin{minipage}{.95\textwidth}
  \fontsize{7pt}{8pt}\selectfont
  $\ast$ Used for cross-referencing image pairs between \texttt{cap.sample.json} and \texttt{cap.ext.sample.json}.\\[1.75pt]
  $\dagger$ See \figref{fig:amt_interface} (a) for the three possible labels. When constructing \texttt{target\_soft}, images labelled as [The same image] is added as \texttt{1.0}, [No differences worth mentioning] is added as \texttt{0.5}, [Images that are too different] is added as \texttt{-1.0}.\\[1.75pt]
  $\ddagger$ See \figref{fig:amt_interface} (b) for the options we provide for AMT workers.\\[1.75pt]
  ${\S}$ Not public for test-split. Instead, see our project website for the test-split evaluation server. \\[1.75pt]
\end{minipage}
  \caption{Data structure as in the data files. For details please refer to our project website.}
  \label{tab:json}
  \end{table*}

  \tabref{tab:json} summarizes information we provide for each image pair.

\end{document}